\documentclass[lettersize,journal]{IEEEtran}
\usepackage{amsmath,amsfonts}
\usepackage{algorithmic}
\usepackage{algorithm}
\usepackage{array}
\usepackage[caption=false,font=normalsize,labelfont=sf,textfont=sf]{subfig}
\usepackage{textcomp}
\usepackage{stfloats}
\usepackage{url}
\usepackage{verbatim}
\usepackage{graphicx}
\usepackage{cite}
\usepackage{bm}
\usepackage{balance}
\usepackage[table]{xcolor}
\usepackage{colortbl}
\usepackage{booktabs}

\begin{document}

\title{A Reconfigurable Tracked Robot for Enhanced Obstacle Traversal Through Movable Articulation Point and Internal Mass Relocation}

\author{Yuki~Uda,~\IEEEmembership{Student~Member,~IEEE,},~Yasutaka~Nakashima,~\IEEEmembership{Member,~IEEE},~Motoji~Yamamoto,~\IEEEmembership{Member,~IEEE},~Ayato~Kanada,~\IEEEmembership{Member,~IEEE}% <-this % stops a space
\thanks{This work was supported by  JSPS KAKENHI Grant Number 26K22451 and the Tateisi Science and Technology Foundation.

Y. Uda, Y. Nakashima, and M. Yamamoto are with the Department
of Mechanical Engineering, Kyushu University, 744 Motooka, Nishi-ku, Fukuoka 819-0395, Japan. 

A. Kanada is with the Graduate School of Informatics and Engineering, The University of Electro-Communications, 1-5-1 Chofugaoka, Chofu-shi, Tokyo, 182-8585, Japan. 
e-mail: (Corresponding author: kanada@uec.ac.jp).}% <-this % stops a space
}

\markboth{Journal of \LaTeX\ Class Files,~Vol.~14, No.~8, August~2021}%
{Shell \MakeLowercase{\textit{et al.}}: A Sample Article Using IEEEtran.cls for IEEE Journals}


\maketitle

\begin{abstract}
Tracked robots are widely used in unstructured environments; however, their obstacle traversal capability is fundamentally limited by a tradeoff between front-end reachability and locomotion stability. This study presents TRASER (Tracked Robot with Articulated Spine for Extended Reach), a reconfigurable tracked robot capable of relocating both its articulation point and internal mass. TRASER employs a tape-spring mechanism that localizes compliance to the bending region while maintaining high stiffness in the remaining body, thereby improving both front-end reachability and center-of-mass (CoM) shifting capability. Geometric and static models are developed to analyze the effects of articulation point and CoM position on step and ditch traversal performances. Experiments demonstrate step traversal, suspended-platform traversal, and ditch traversal of 74\%, 66\%, and 59\% of the robot body length, respectively. To the best of our knowledge, these results represent the highest reported obstacle traversal capabilities among tracked mobile robots.
\end{abstract}

\begin{IEEEkeywords}
Tracked robot, reconfigurable robot, mobile robot, robot mechanism.
\end{IEEEkeywords}

\section{Introduction}

\begin{figure*}[!t]
\centering
\includegraphics[width=0.90\linewidth]{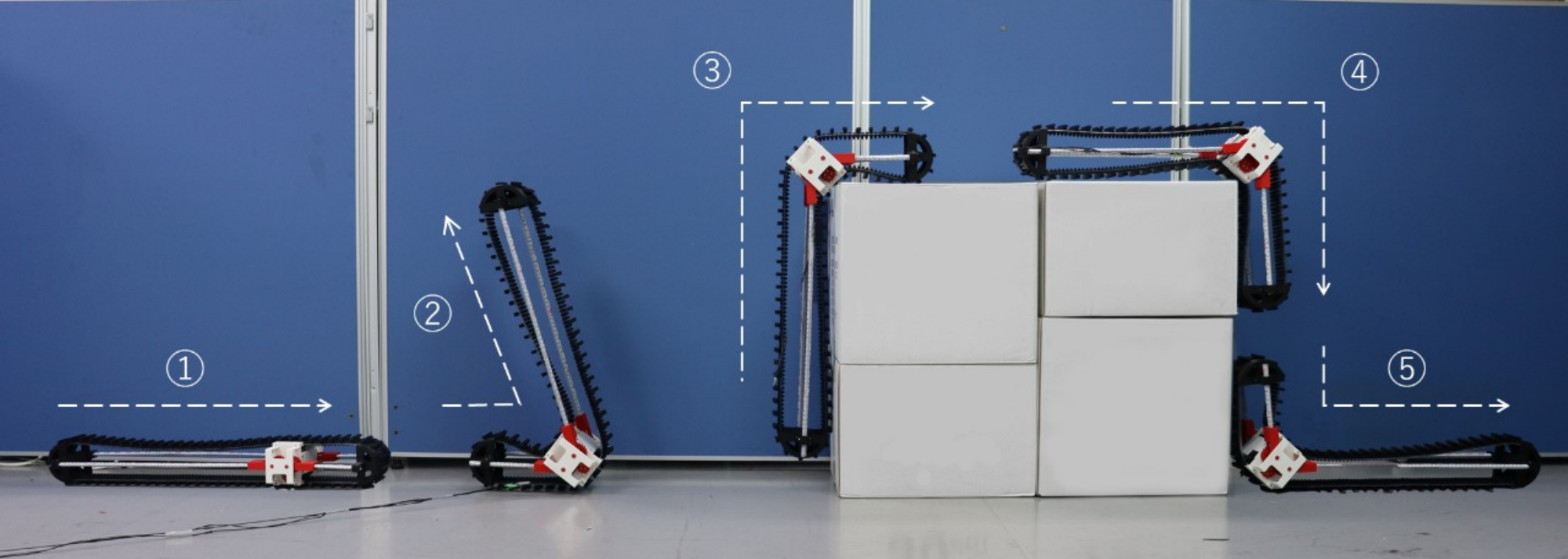}
\caption{Overview of TRASER. By relocating the bending unit along the body, the robot achieves both high front-end reachability and CoM relocation, enabling traversal of large obstacles.}
\label{samune}
\end{figure*}

\IEEEPARstart{T}{racked} robots provide high traction and locomotion stability owing to their large ground contact area, making them well suited for traversing unstructured environments such as gravel, damaged roads, and rubble~\cite{GARCIA2024,Tong2024SARReview,6386301}. Consequently, they have been widely used in disaster response, infrastructure inspection, and construction. However, these environments often contain large steps, gaps, and debris that restrict robot mobility. Improving obstacle traversal capability therefore remains one of the central challenges in tracked robot research.

For high-step traversal, sufficient driving force alone is not enough. Before traction can be generated, the front end of the track must first reach the upper surface of the obstacle to obtain the required support reaction force. Consequently, the maximum traversable obstacle height is fundamentally limited by the front-end reachability of the track.

To improve front-end reachability, numerous tracked robots equipped with flippers/sub-tracks~\cite{quince,kenaf,6kurora}, variable-geometry bodies~\cite{kahen1,kahen2,Xu2024,Kinugasa2016}, track-linkage mechanisms~\cite{Lim2025}, and articulated mechanisms~\cite{souryu,tyokuretuEN} have been proposed. These mechanisms increase traversal adaptability by changing the body configuration. Nevertheless, the achievable front-end reachability is still fundamentally constrained by the articulation-point location and the center-of-mass (CoM) position.

Increasing front-end reachability generally requires the robot to bend about a support point near the rear of the body. However, moving the articulation point rearward shifts the CoM toward the edge of the support polygon, increasing the risk of tip-over. Conversely, shifting the CoM rearward improves stability but makes it more difficult to transfer the robot body onto the obstacle after initial contact. Consequently, tracked robots inherently face a tradeoff between front-end reachability and locomotion stability.

Several approaches have been proposed to mitigate the tradeoff between front-end reachability and locomotion stability. Internal mass-shifting mechanisms relocate the robot's CoM to improve stability during obstacle traversal~\cite{app12010525,Hirose1992TAQT}, whereas tail-assisted structures provide an additional support point to suppress tip-over~\cite{Chiu2005,Seo2013FlipBot,Kececi2009DesignAP,Gao2017WheelTrack}. Another strategy is to exploit reaction forces from obstacle walls to lift the track front end~\cite{Soltanzadeh2011COG}. Although effective, these approaches typically increase mechanical complexity, enlarge the robot body, rely on specific obstacle geometries, or produce direction-dependent locomotion performance.

More recently, reconfigurable tracked robots capable of relocating the articulation point have been proposed~\cite{8930917}. These robots improve adaptability to different obstacle geometries; however, because their mass distribution remains unchanged, the reachability--stability tradeoff is fundamentally preserved.

In our previous work, we proposed a tracked robot with an internally movable motor unit that enabled arbitrary-location articulation using a mechanically simple structure~\cite{uda2025crawler}. However, the compliant body suffered from significant sagging during large bending, reducing front-end reachability. Moreover, because the movable motor unit represented only a small fraction of the total robot mass, its CoM-shifting capability was limited. These observations suggest that superior obstacle traversal requires both high front-end reachability and substantial CoM relocation, while maintaining a compact and mechanically simple design.

To address this challenge, we employ tape springs, which have been widely used in deployable robotic and space structures because they provide high stiffness while allowing localized compliant deformation~\cite{doi:10.1126/sciadv.adt5905,ding2022planar,chen2024locomotion,osele2022lightweight,seffen1999deployment}. When buckled, a tape spring forms a localized compliant region that can be translated along its length, effectively creating an articulation point while preserving high stiffness throughout the remaining structure.

Based on this concept, we present TRASER, an enhanced reconfigurable tracked robot that combines a movable internal motor unit with paired tape springs (Fig.~\ref{samune}). TRASER localizes compliance exclusively to the bending region, suppressing body sagging even under large bending configurations. Furthermore, the high-stiffness structure eliminates additional supporting components required by compliant-body designs, reducing the overall robot weight and increasing the relative movable mass of the bending unit. Consequently, TRASER simultaneously improves front-end reachability and CoM relocation capability, which overcomes key limitations of existing tracked robots.

The main contributions of this study are summarized as follows:

\begin{enumerate}
\item We propose a reconfigurable tracked robot capable of arbitrary-location articulation and substantial internal mass relocation using a simple mechanical architecture.

\item We develop analytical models to clarify how articulation location and CoM position influence step traversal and ditch traversal performance, providing design guidelines for obstacle traversal.

\item Experiments demonstrate step traversal, suspended-platform traversal, and ditch traversal of 74\%, 66\%, and 59\% of the robot body length, respectively. To the best of our knowledge, these represent the highest obstacle traversal capabilities reported for tracked mobile robots.
\end{enumerate}

\section{Robot Design}

\subsection{Design Concept}
TRASER aims to improve obstacle traversal by actively adjusting both the articulation point and the CoM position. To achieve this, a movable bending unit that translates along the robot body is introduced. Because it contains the major actuators of the robot, relocating the unit shifts both the articulation-point location and the CoM.

Maintaining high body stiffness is equally important because body sagging reduces front-end reachability during large bending. Therefore, the robot employs paired tape springs that localize compliance only to the bending region while preserving high stiffness throughout the remaining body.

TRASER consists of three main components: a tape-spring spine, a track belt, and a movable bending unit that controls both articulation-point location and CoM position.

\begin{figure*}[!t]
\centering
\includegraphics[width=0.90\linewidth]{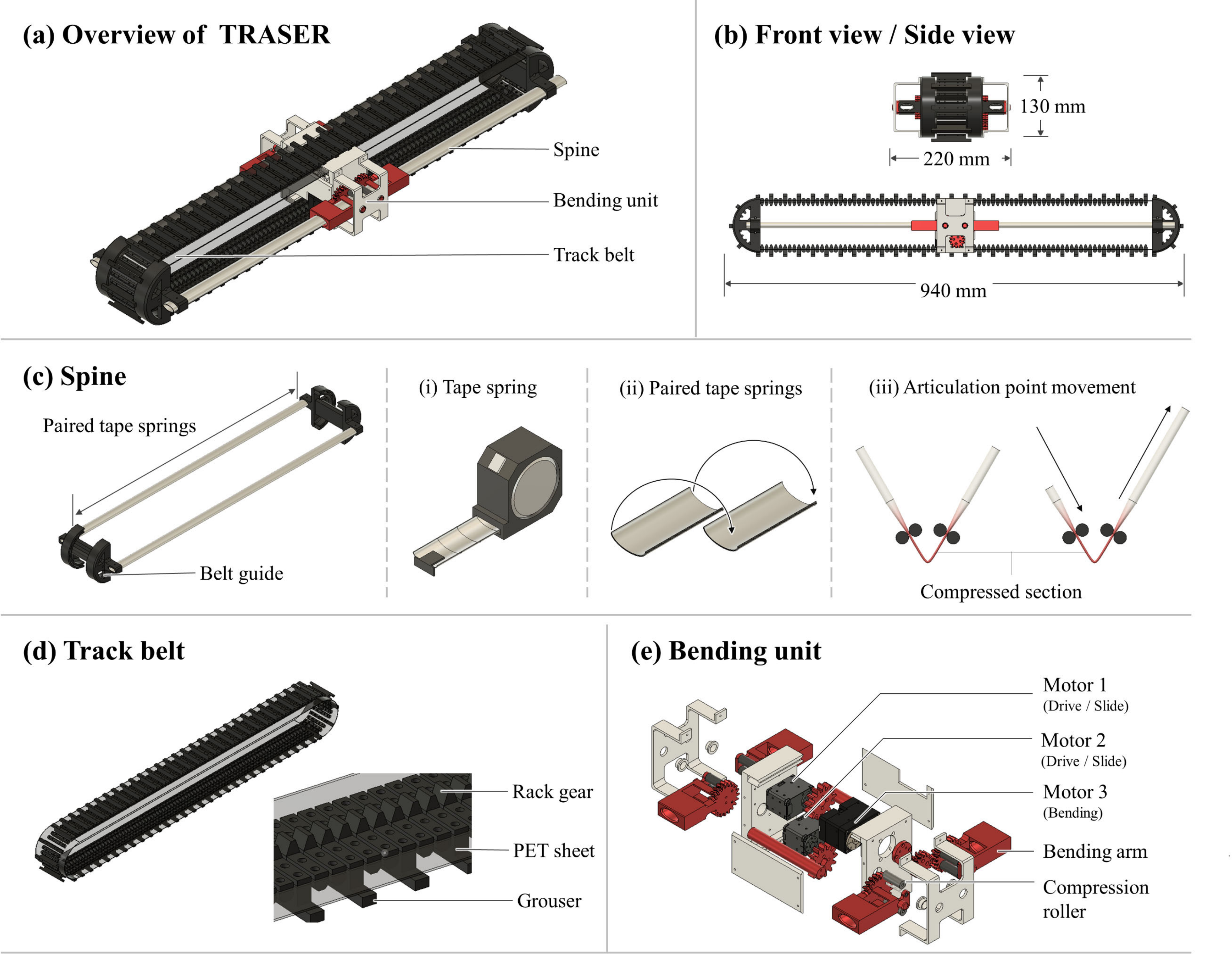}
\caption{Mechanical design of TRASER. (a) Overall appearance. (b) Dimensions. (c) Spine module composed of paired tape springs and belt guides. (d) Track belt. (e) Exploded view of the movable bending unit.}
\label{robot_design}
\end{figure*}

\subsection{Mechanical Design}
Fig.~\ref{robot_design} illustrates TRASER. The robot has an overall length of 940 mm, a height of 130 mm, a width of 220 mm, and a mass of 2.5 kg. The following subsections describe the three main components of the robot: the spine, the track belt, and the movable bending unit.

\subsubsection{Spine}
The spine consists of paired tape springs and belt guides. Since a single tape spring exhibits asymmetric bending stiffness, two tape springs are bonded with their concave surfaces facing each other to form a symmetric closed section with nearly identical stiffness in both bending directions. Two such spine modules are installed on the left and right sides of the robot. Semicircular belt guides retain the track belt and prevent derailment caused by belt slack during large body deformation.

\subsubsection{Track Belt}
The track belt is fabricated from a rolled polymer sheet. Rack gears are formed on the inner surface of the belt to engage with the drive motors, while grousers are attached to the outer surface to generate traction against the ground and obstacles. The grousers are coated with rubber to further improve grip. Polyethylene terephthalate (PET) is selected as the belt material because of its high impact resistance and low friction coefficient against the belt guides.

\subsubsection{Movable Bending Unit}
The movable bending unit contains two drive motors (Dynamixel XC430-240-T, Robotis) and one bending motor (Dynamixel XM540-270-T, Robotis). The drive motors engage the rack gears on the track belt to realize both locomotion and unit translation. During translation, the tape springs function as guide rails, allowing accurate positioning independent of surface friction. Because it houses the major actuators, the unit has a mass of $m_u = 0.878\,\mathrm{kg}$, accounting for approximately 35\% of the total robot mass.

The bending motor drives a bending arm that locally flattens the paired tape springs using compression rollers, thereby generating a movable compliant region. Because the unit is not fixed to either the body or the track belt, it can translate freely along the robot body over a range of $630\,\mathrm{mm}$, enabling arbitrary-location articulation. This translation shifts the center of mass by about $222\,\mathrm{mm}$, corresponding to approximately 24\% of the robot body length. Belt guides mounted on the unit maintain belt engagement throughout deformation.

\subsection{Operating Principle}
\begin{figure}[!t]
\centering
\includegraphics[width=0.90\linewidth]{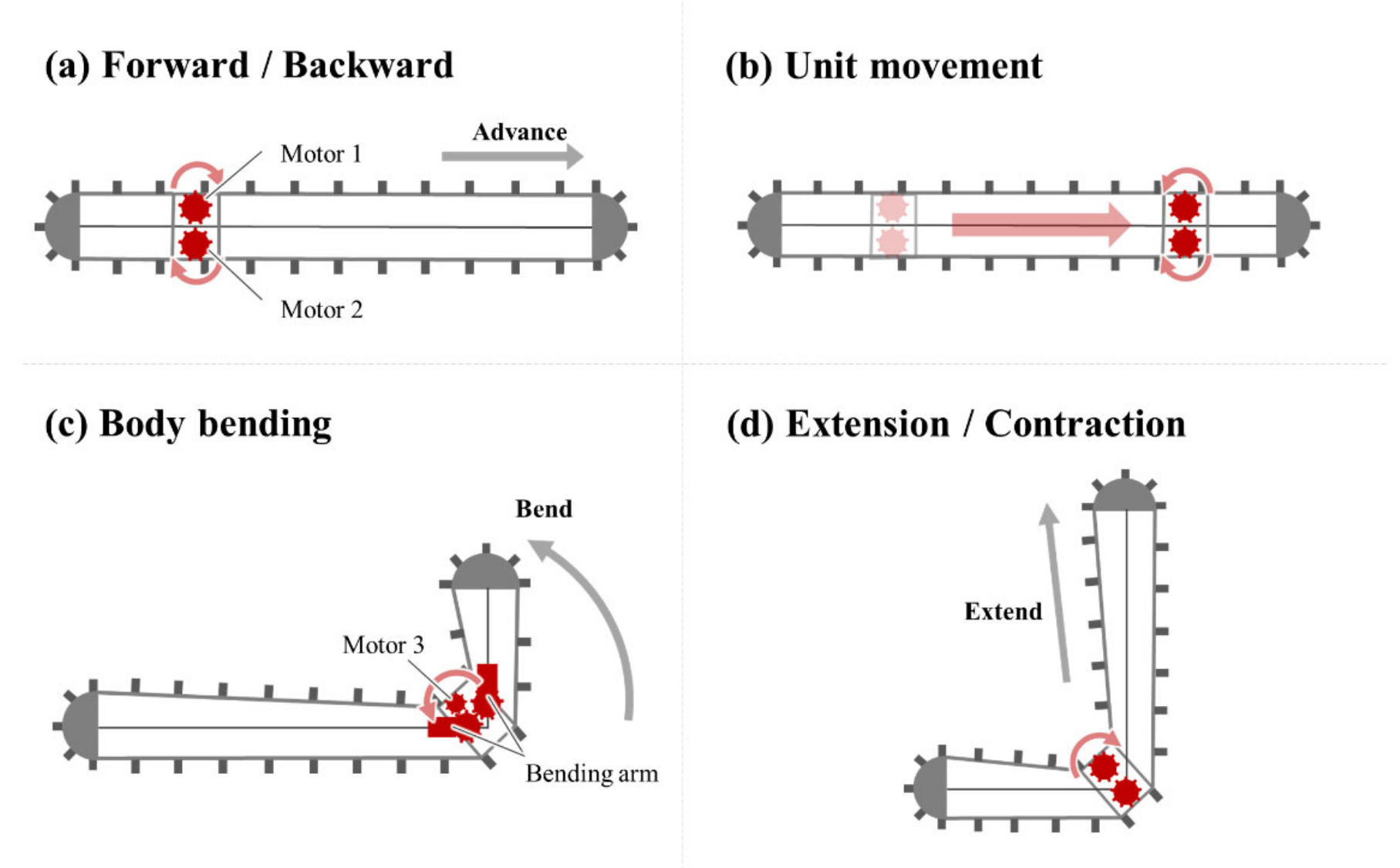}
\caption{Four fundamental motions of TRASER. (a) Tracked locomotion. (b) Translation of the bending unit. (c) Body bending. (d) Body extension and contraction.}
\label{robot_motion}
\end{figure}

The operating principle of TRASER is illustrated in Fig.~\ref{robot_motion}. The robot is actuated by three motors, enabling four fundamental operation modes: tracked locomotion, bending-unit translation, body bending, and body extension/contraction. The two motors driving the track belt are referred to as Motors 1 and 2, whereas the motor actuating the bending arm is referred to as Motor 3.

When Motors 1 and 2 rotate synchronously in the same direction (Fig.~\ref{robot_motion}(a)), the track belt circulates around the robot body, generating forward or backward locomotion in the same manner as a conventional tracked robot. In contrast, when the two motors rotate in opposite directions (Fig.~\ref{robot_motion}(b)), the belt motion is canceled and the bending unit translates along the body. Because the articulation point is generated at the position of the bending unit, this operation relocates the articulation point without changing the overall body configuration.

Body bending is generated by Motor 3 (Fig.~\ref{robot_motion}(c)). Through a gear transmission, the bending arm locally flattens the paired tape springs using compression rollers, creating a compliant bending region while the remaining body retains its high stiffness. Since the bending unit can be positioned anywhere along the robot body, an articulation point can be generated at an arbitrary location.

Finally, actuating only Motor 1 produces body extension and contraction (Fig.~\ref{robot_motion}(d)). The rear side of the track belt is reeled in while the front side is released, causing the unsupported body section to extend or contract and gradually translating the bending unit toward the rear. As a result, the front end is raised while the center of mass shifts rearward, improving stability during obstacle traversal.

By combining these four operation modes, the robot can continuously adjust its body configuration, articulation point, and CoM position according to the surrounding environment, enabling versatile traversal over obstacles with various geometries. 

\section{Obstacle Traversal Strategy}
This section presents obstacle traversal strategies for three challenging obstacle scenarios: vertical steps, suspended-platforms, and ditches. Each strategy is realized by combining the four operation modes introduced in Section II.

\subsection{Step Traversal Strategy}
\begin{figure}[!t]
\centering
\includegraphics[width=0.90\linewidth]{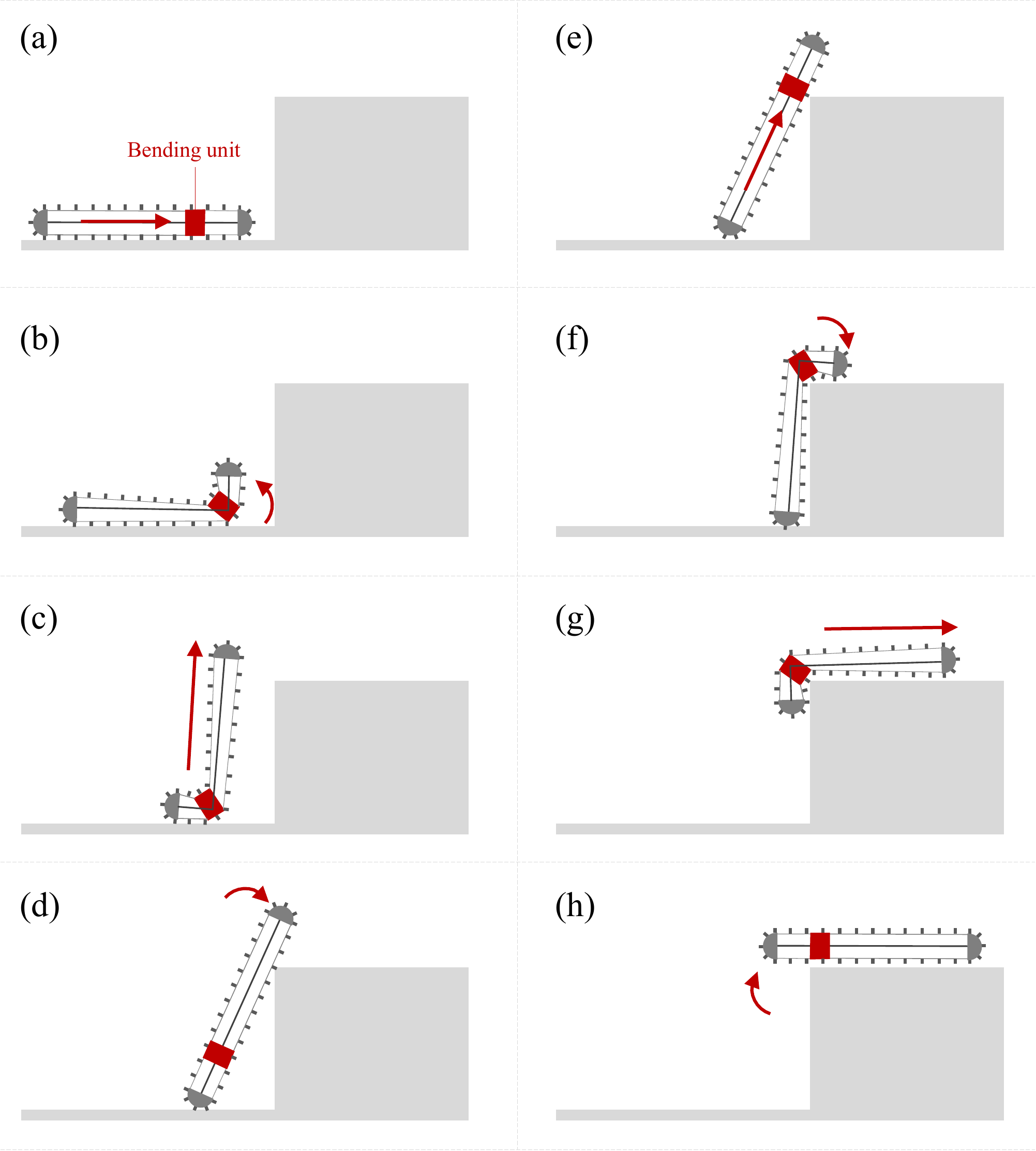}
\caption{Step traversal sequence. The robot reaches the upper surface by combining movable articulation with wall-assisted stabilization.}
\label{step_str}
\end{figure}

The proposed step traversal sequence is illustrated in Fig.~\ref{step_str}. The robot first translates the bending unit toward the front of the body (a) and bends the body to raise the front end above the upper surface of the step (b,c). After establishing contact with the upper surface (d), the bending unit is translated onto the step and the body is bent again to maintain contact with both the wall and the upper surface (e,f). Finally, the robot advances onto the step and restores its straight configuration (g,h).

During the latter stage of the maneuver, the reaction force generated by the wall suppresses the tip-over moment, allowing the robot to exploit its high front-end reachability while maintaining stability.

\subsection{Suspended-Platform Traversal Strategy}
\begin{figure}[!t]
\centering
\includegraphics[width=0.90\linewidth]{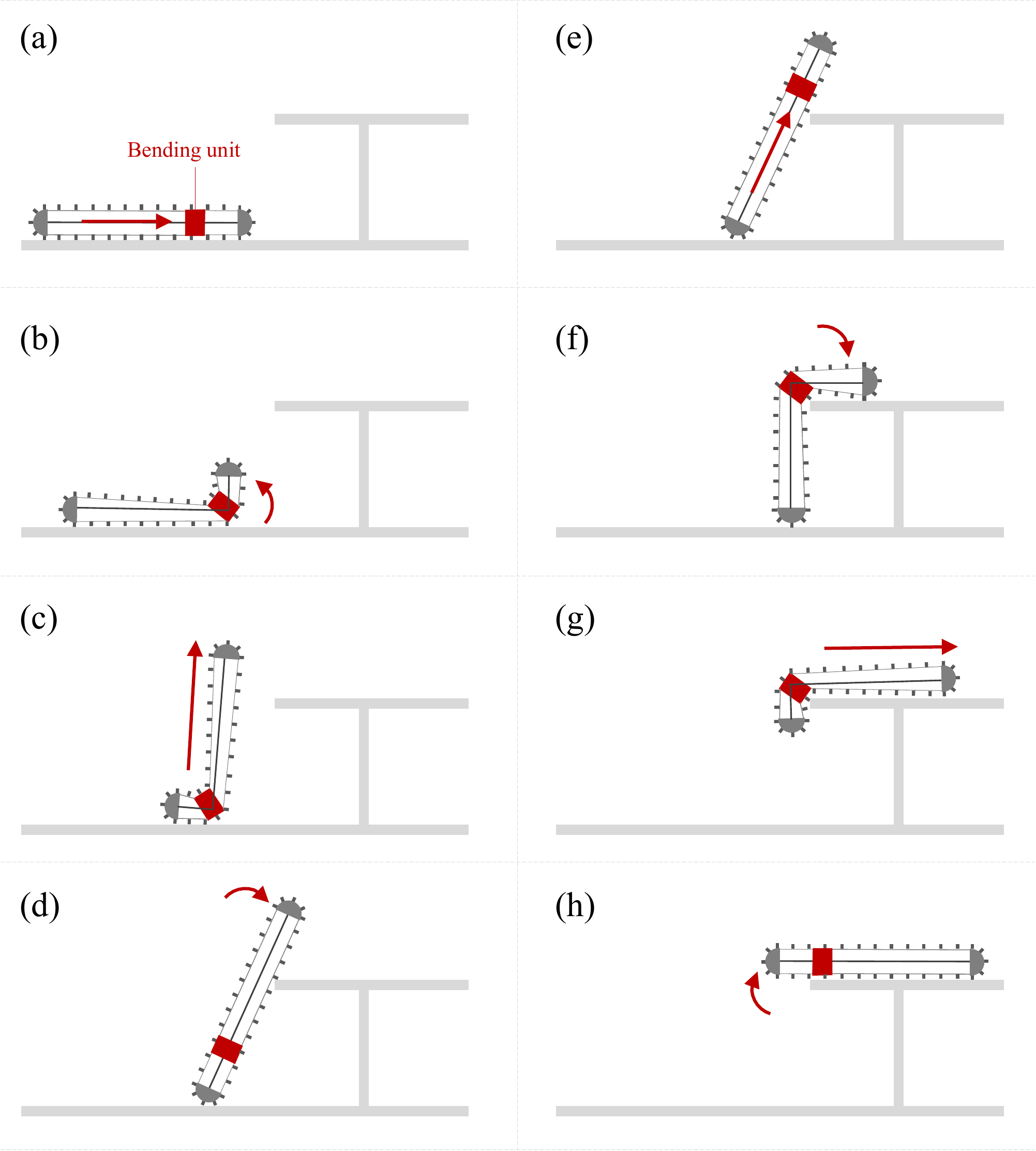}
\caption{Suspended-platform traversal sequence. The robot traverses the platform without relying on wall reaction by actively relocating its center of mass.}
\label{tenban_str}
\end{figure}

The suspended-platform traversal strategy is illustrated in Fig.~\ref{tenban_str}. The robot first translates the bending unit toward the front of the body (a) and bends the body to raise the front end above the platform (b,c). After establishing contact with the upper surface (d), the bending unit is translated further forward until it is positioned above the platform (e). The body is then bent again while maintaining contact with the platform, allowing the front body to advance onto the upper surface while the rear body remains vertically supported (f). Finally, the robot moves its front body completely onto the platform and restores its straight configuration (g,h).

Unlike the step traversal strategy, no wall reaction force is available to suppress the tip-over moment after the front end reaches the platform. Therefore, the robot must actively relocate its center of mass onto the platform before the rear end leaves the ground. This is achieved by coordinating arbitrary-location articulation with substantial CoM relocation, enabling stable traversal of elevated structures without relying on vertical supporting surfaces.

\subsection{Ditch Traversal Strategy}
The robot first translates the bending unit toward the front of the body (a), relocating the articulation point while keeping the center of mass near the rear of the robot. The body is then bent to lift the front end (b), allowing it to reach the opposite edge (c). After establishing contact with the far side by straightening the body (d), the bending unit is translated forward (e,f), shifting the center of mass onto the opposite side before the rear end leaves the ground (g). Finally, the robot completely crosses the ditch (h). By maintaining a rearward center of mass during opposite-edge reaching and subsequently shifting it forward before rear-end lift-off, the robot avoids falling into the ditch while traversing gaps substantially wider than those negotiable by conventional tracked robots.

\begin{figure}[!t]
\centering
\includegraphics[width=0.90\linewidth]{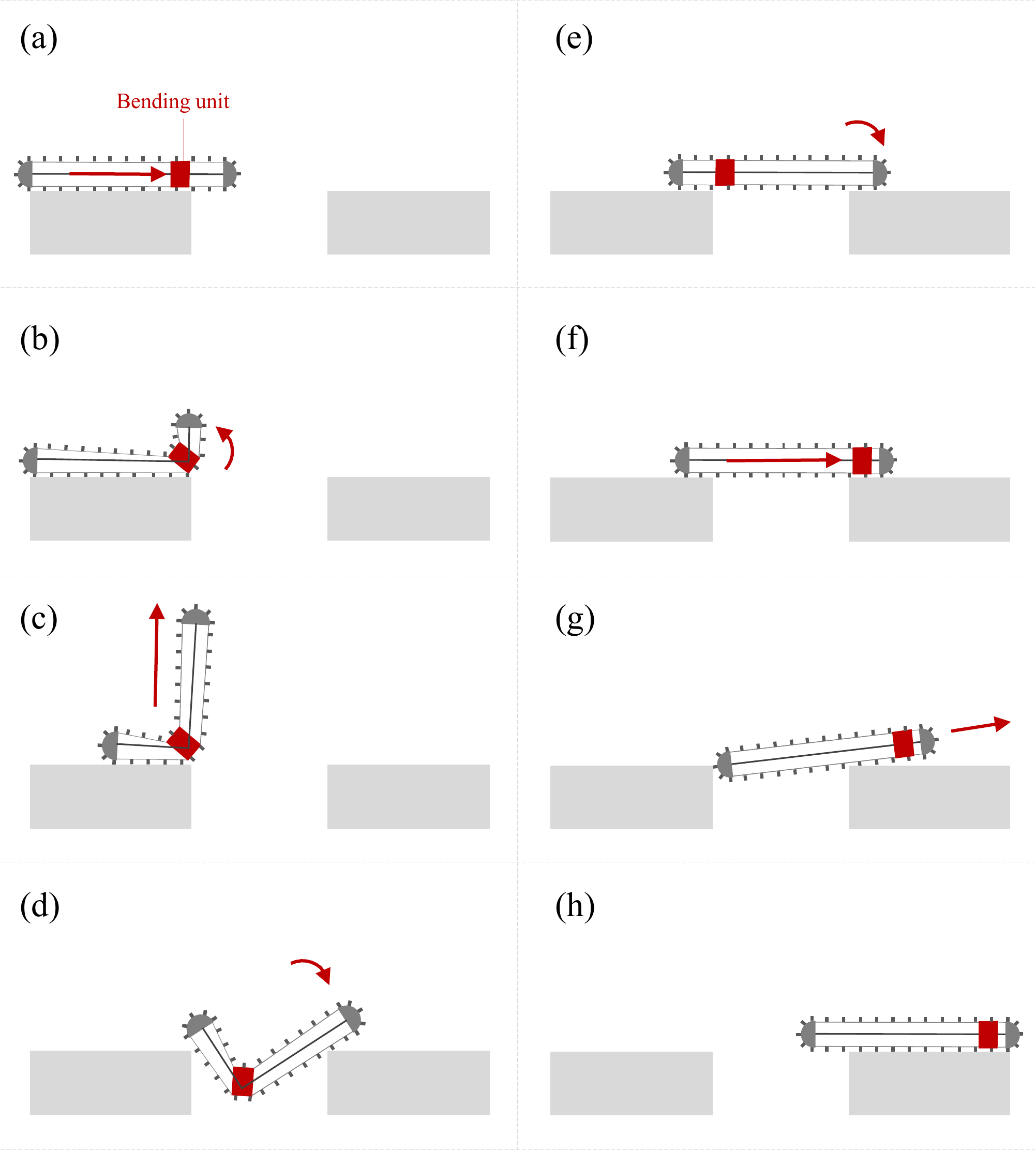}
\caption{Ditch traversal sequence. The robot first reaches the opposite edge and then shifts its center of mass forward to achieve stable gap traversal.}
\label{beyond_str}
\end{figure}

\section{Modeling}
This section develops analytical models to evaluate the feasibility of the proposed mechanism and its obstacle traversal performance.

First, the variation in belt path length caused by body bending is analyzed to verify the feasibility of the belt mechanism. Next, the robot's center of mass (CoM) is derived to determine the combinations of bending angle and unit position that prevent tip-over during front-end lifting. Finally, analytical models are developed for step traversal, suspended-platform traversal, and ditch traversal to derive the theoretical traversal limits for each obstacle.

Throughout this section, the unsupported portions of the track belt are assumed to remain taut due to belt tension. In addition, the track belt is assumed to be sufficiently flexible such that bending deformation does not affect the analytical results. The physical parameters of the prototype robot used throughout the following analyses are summarized in Table~\ref{tab:geometric_parameters}.
 
\begin{table}[!t] \centering \caption{Physical Parameters of the Prototype Robot} \label{tab:geometric_parameters} \begin{tabular}{c l c} \toprule \textbf{Symbol} & \textbf{Description} & \textbf{Value} \\ \midrule $L$ & Total length of the tape spring (body) & 810~mm \\ $r$ & Arc radius of the belt guide & 50~mm \\ $a$ & Distance between gears in the bending unit & 40~mm \\ $b$ & Distance from gear center to belt guide end & 17.5~mm \\ $h_g$ & Height of the grouser & 9.4~mm \\ $\rho_t$ & Linear mass density of the track belt & 0.568~g/mm \\ $\rho_c$ & Linear mass density of the tape spring & 0.345~g/mm \\ $m_r$ & Concentrated mass of a belt guide & 109~g \\ $m_u$ & Mass of the movable bending unit & 878~g \\ $M$ & Total mass of the robot & 2496~g \\ \bottomrule \end{tabular} \end{table}

\subsection{Belt Path Length Variation with Articulation Point}
An articulation point changes the geometry of the track belt path. Excessive variation in the belt path length may cause belt slack or excessive belt tension, leading to unstable power transmission or mechanical interference. Therefore, evaluating the belt-path variation is essential for verifying the feasibility of the proposed mechanism.

To quantify this effect, the \emph{belt path length variation ratio} is defined as the change in the total belt path length normalized by the belt path length in the initial straight configuration. Conventional tracked robots employing tensioned belts typically require this variation to remain within approximately 1\% to prevent excessive belt deformation. In contrast, TRASER employs belt guides together with unsupported belt sections that absorb moderate path-length variations. Therefore, an allowable variation ratio of 5\% is adopted in this study.

\begin{figure}[!t]
\centering
\includegraphics[width=0.90\linewidth]{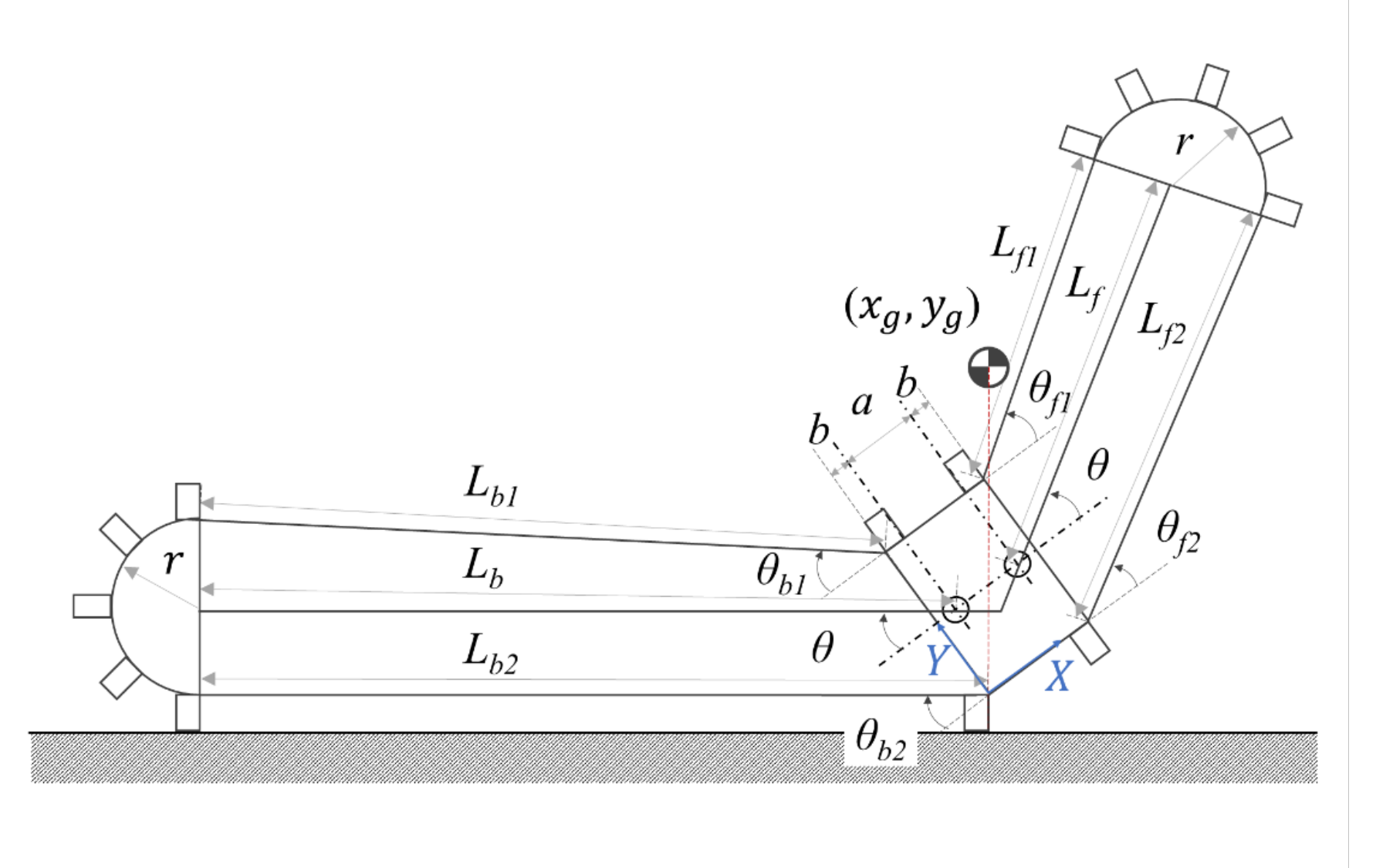}
\caption{Geometric model for deriving the belt-path variation rate from the bending angle $\theta$ and unit position $L_b$.}
\label{zentyou_mdr}
\end{figure}

The belt-path variation is evaluated using the geometric model shown in Fig.~\ref{zentyou_mdr}. The model treats the bending angle $\theta$ and rear body length $L_b$ as independent variables, from which the belt path length for each articulated configuration is derived. Here, $\theta$ denotes the rotation angle of each arm; therefore, the overall robot bending angle is $2\theta$. The obtained belt path length is then compared with that of the initial straight configuration to calculate the variation ratio.

Here, $L_b$ and $L_f$ denote the rear and front tape-spring lengths, and $L_{b1}$, $L_{b2}$, $L_{f1}$, $L_{f2}$ denote the corresponding upper and lower belt path lengths, with belt deflection angles $\theta_{b1}$, $\theta_{b2}$, $\theta_{f1}$, and $\theta_{f2}$.

Based on the geometry shown in Fig.~\ref{zentyou_mdr}, the lengths of the four straight belt segments are obtained as
\begin{align}
    L_{b1} &= \sqrt{(L_b \cos\theta - b - r\sin\theta)^2 + (L_b \sin\theta - r + r\cos\theta)^2} \\
    L_{b2} &= \sqrt{(L_b \cos\theta - b + r\sin\theta)^2 + (L_b \sin\theta + r - r\cos\theta)^2} \\
    L_{f1} &= \sqrt{(L_f \cos\theta - b - r\sin\theta)^2 + (L_f \sin\theta - r + r\cos\theta)^2} \\
    L_{f2} &= \sqrt{(L_f \cos\theta - b + r\sin\theta)^2 + (L_f \sin\theta + r - r\cos\theta)^2}
\end{align}
where the front tape-spring length $L_f$ is expressed as
\begin{equation}
    L_f = L - L_b - \frac{a}{\cos\theta}
\end{equation}
from the total body length $L$ and the distance $a$ between the two drive gears. The corresponding belt deflection angles are given by
\begin{align}
    \theta_{b1} &= \operatorname{atan2}\!\left(L_b \sin\theta - r + r\cos\theta,\; L_b \cos\theta - b - r\sin\theta\right) \\
    \theta_{b2} &= \operatorname{atan2}\!\left(L_b \sin\theta + r - r\cos\theta,\; L_b \cos\theta - b + r\sin\theta\right) \\
    \theta_{f1} &= \operatorname{atan2}\!\left(L_f \sin\theta - r + r\cos\theta,\; L_f \cos\theta - b - r\sin\theta\right) \\
    \theta_{f2} &= \operatorname{atan2}\!\left(L_f \sin\theta + r - r\cos\theta,\; L_f \cos\theta - b + r\sin\theta\right)
\end{align}
Using these geometric relationships, the belt-path variation ratio is defined as
\begin{equation}
    f(\theta, L_b) = \frac{(L_{f1} + L_{f2} + L_{b1} + L_{b2}) - 2(L - a - 2b)}{2(L + \pi r)}
\end{equation}
where the initial belt path length in the straight configuration is $2(L+\pi r)$.

\begin{figure}[!t]
\centering
\includegraphics[width=0.90\linewidth]{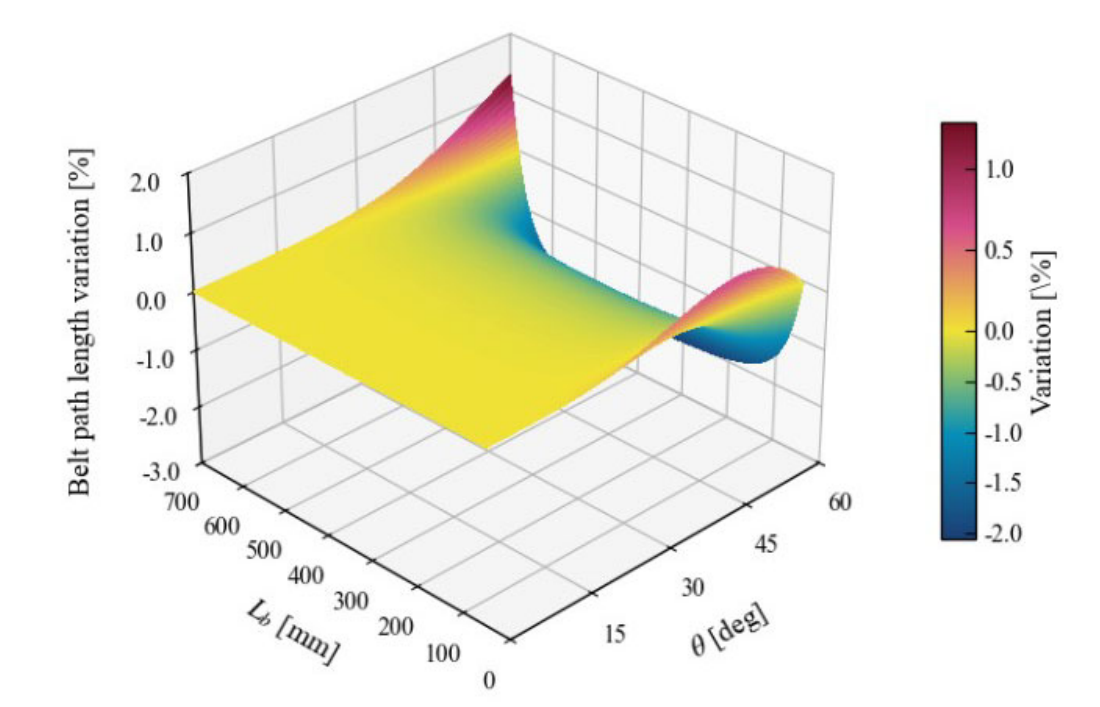}
\caption{Calculated belt-path length variation as functions of the bending angle $\theta$ and bending-unit position $L_b$. The variation remains below $3\%$ throughout the operating range.}
\label{kyokumen_grp}
\end{figure}

Fig.~\ref{kyokumen_grp} shows the calculated belt-path variation ratio as a function of the bending angle $\theta$ and the rear tape-spring length $L_b$ using the parameters of the prototype robot. The maximum variation ratio remains below $3\%$ over the entire range of bending angles and articulation point, satisfying the design criterion of $5\%$. Therefore, the proposed mechanism maintains an almost constant belt path length throughout reconfiguration, confirming the feasibility of the belt transmission mechanism without requiring an active belt-tensioning system.

\subsection{CoM Analysis and Stability Condition}

This subsection derives the center of mass (CoM) of TRASER and determines the conditions under which the robot remains statically stable while lifting its front body. During large bending motions, the CoM shifts toward the front of the support polygon, increasing the risk of tip-over. Therefore, the relationship between the bending angle $\theta$ and the rear body length $L_b$ is analyzed to identify the feasible operating region. The derived CoM model is also used in the subsequent analyses of obstacle traversal.

Static stability is maintained as long as the projection of the CoM onto the ground contact surface remains inside the support polygon.

The robot is modeled as a collection of uniformly distributed structural members and concentrated masses corresponding to the belt guides and the movable bending unit. Based on this mass distribution, the overall CoM in the global coordinate system, $(x_g, y_g)$, is obtained as

\begin{align}
    x_g &= \frac{1}{M} \left[- \frac{\rho_t L_{b1}^2}{2}\cos\theta_{b1} - \frac{\rho_t L_{b2}^2}{2}\cos\theta_{b2} \right. \nonumber \\
    & \quad - \left(m_r + \rho_t \pi r\right)\left\{\left(L_b + \frac{2r}{\pi}\right)\cos\theta - b\right\} \nonumber \\
    & \quad - \rho_c L_b \left(\frac{L_b}{2}\cos\theta - b\right) \nonumber \\
    & \quad + \left\{ m_u + 2\rho_t(a+2b) + \frac{\rho_c a}{\cos\theta} \right\}\frac{a+2b}{2} \nonumber \\
    & \quad + \rho_t L_{f1}\left(\frac{L_{f1}}{2}\cos\theta_{f1} + a + 2b\right) \nonumber \\
    & \quad + \rho_t L_{f2}\left(\frac{L_{f2}}{2}\cos\theta_{f2} + a + 2b\right) \nonumber \\
    & \quad + \rho_c L_f \left(\frac{L_f}{2}\cos\theta + a + b\right) \nonumber \\
    & \quad \left. + \left(m_r + \rho_t \pi r\right)\left\{\left(L_f + \frac{2r}{\pi}\right)\cos\theta + a + b\right\} \right]
\end{align}

\begin{align}
    y_g &= \frac{1}{M} \left[ \rho_t L_{b1}\left(2r + \frac{L_{b1}}{2}\sin\theta_{b1}\right) \right. \nonumber \\
    & \quad + \frac{\rho_t L_{b2}^2}{2}\sin\theta_{b2} + \rho_c L_b \left(\frac{L_b}{2}\sin\theta + r\right) \nonumber \\
    & \quad + \left(m_r + \rho_t \pi r\right)\left\{\left(L_b + \frac{2r}{\pi}\right)\sin\theta + r\right\} \nonumber \\
    & \quad + \left\{ m_u + 2\rho_t(a+2b) + \frac{\rho_c a}{\cos\theta} \right\}r \nonumber \\
    & \quad + \rho_t L_{f1}\left(2r + \frac{L_{f1}}{2}\sin\theta_{f1}\right) + \frac{\rho_t L_{f2}^2}{2}\sin\theta_{f2} \nonumber \\
    & \quad + \rho_c L_f \left(\frac{L_f}{2}\sin\theta + r\right) \nonumber \\
    & \quad \left. + \left(m_r + \rho_t \pi r\right)\left\{\left(L_f + \frac{2r}{\pi}\right)\sin\theta + r\right\} \right]
\end{align}

The tip-over boundary is reached when the CoM projected onto the ground contact surface coincides with the edge of the support polygon, which yields \begin{equation} x_g \cos\theta_{b2} - y_g \sin\theta_{b2} = 0. \end{equation}

\begin{figure}[!t]
\centering
\includegraphics[width=0.90\linewidth]{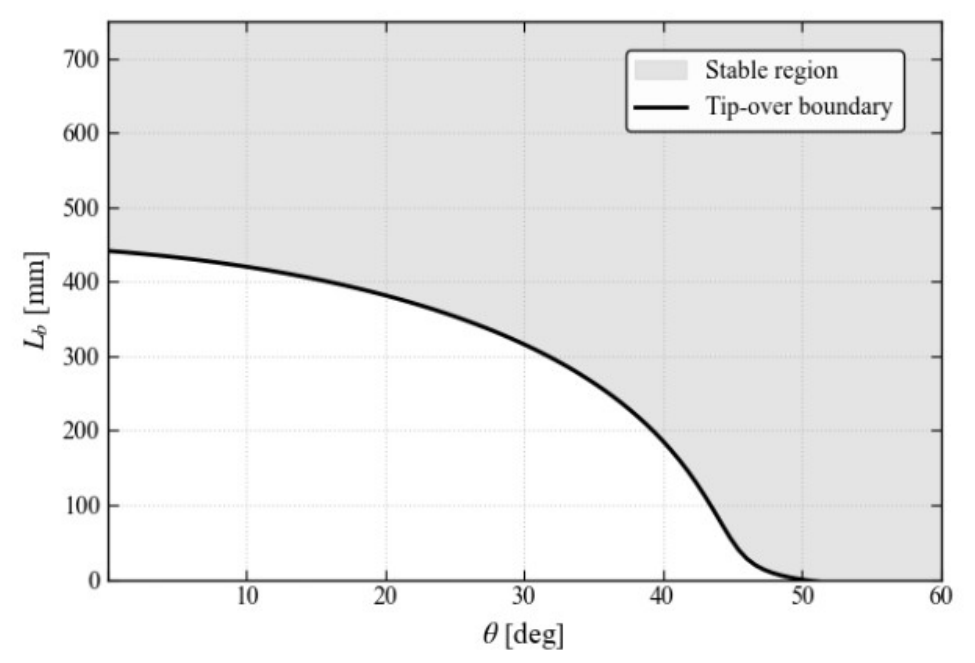}
\caption{Static stability map showing feasible combinations of bending angle $\theta$ and bending-unit position $L_b$. The shaded region satisfies the tip-over prevention condition.}
\label{zentyou_grp}
\end{figure}

Fig.~\ref{zentyou_grp} shows the stability boundary obtained by numerically solving the above equation for various combinations of $\theta$ and $L_b$. The stable region corresponds to the parameter combinations that satisfy the static stability condition.

The analysis indicates that increasing the bending angle shifts the CoM toward the front of the support polygon, thereby requiring a larger rear body length to maintain stability. Consequently, the robot should first increase the bending angle while keeping the bending unit near the front of the body. After the front end has been sufficiently elevated, the bending unit can be translated rearward to relocate the CoM toward the rear, allowing further body bending without tip-over. This stability map provides a theoretical guideline for selecting feasible bending-unit positions and body configurations during obstacle traversal.

\subsection{Step Traversal Model} 
This subsection derives the theoretical maximum step height, $H_{\mathrm{lim}}^{\mathrm{step}}$, of TRASER based on static equilibrium analysis.

As illustrated in Fig.~\ref{step_str}, successful step traversal requires satisfying two conditions. The first condition is that the front end geometrically reaches the upper surface of the step. The second condition is that no slipping occurs at the instant when the rear end leaves the ground during the transition from Fig.~\ref{step_str}(f) to (g).

\begin{figure}[!t]
\centering
\includegraphics[width=0.90\linewidth]{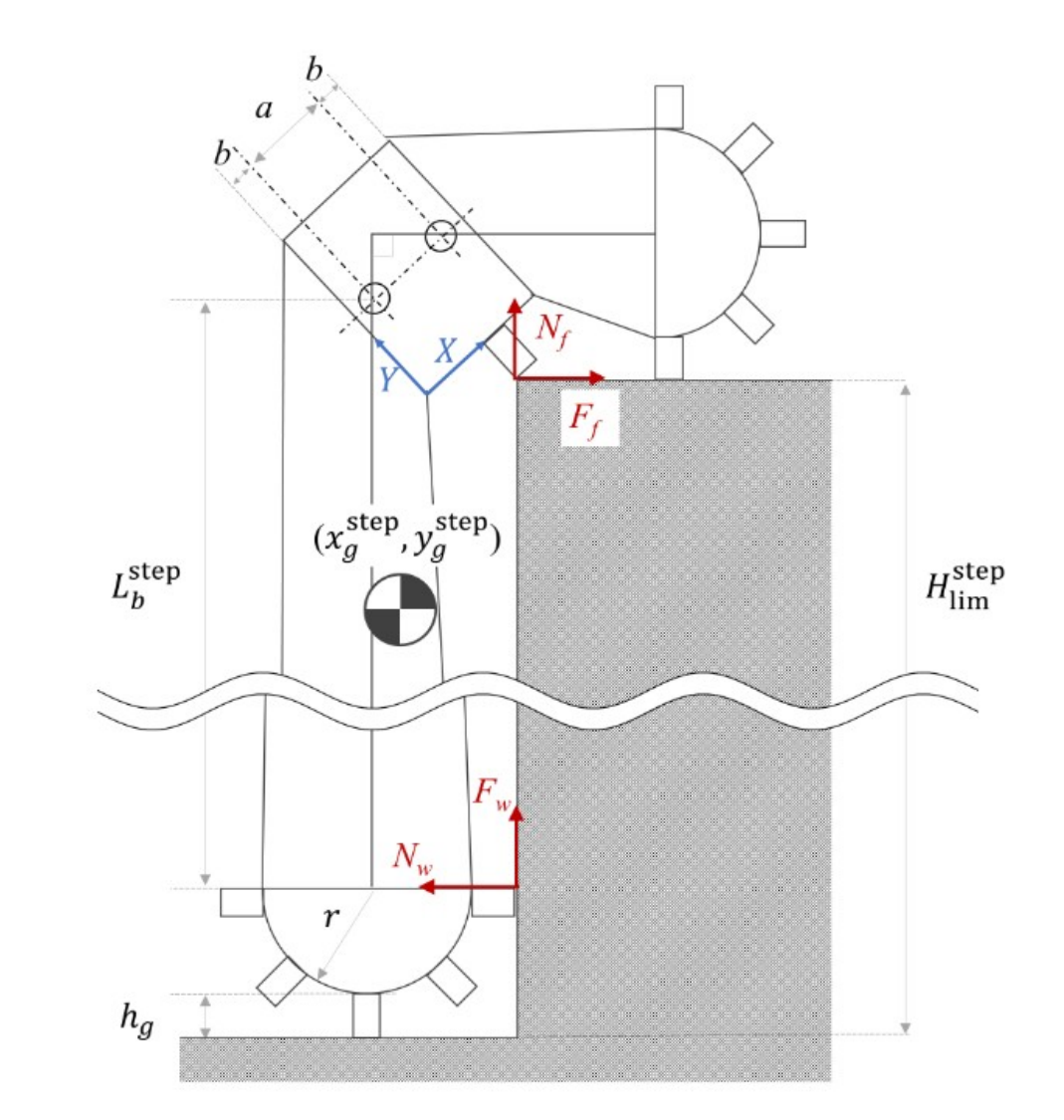}
\caption{Free-body diagram for step traversal analysis. The wall reaction force contributes to preventing tip-over during traversal.}
\label{step_mdr}
\end{figure}

To evaluate these conditions, the model shown in Fig.~\ref{step_mdr} is employed. In this model, the robot is assumed to be in a bent configuration with the bending unit placed at its foremost position. For analytical simplicity, the bending angle is approximated as $\theta = -45^\circ$; the actual angle is slightly shallower because the grouser contacts the wall before the body reaches the ideal right-angle configuration, but this deviation has a negligible effect on the analytical results. The rear body length in this configuration is denoted by $L_b^{\mathrm{step}}$, and the CoM position is then given by (11) and (12) as $(x_g^{\mathrm{step}}, y_g^{\mathrm{step}})$. The normal force, friction force, and friction coefficient at the wall and at the upper surface are denoted by $(N_w, F_w, \mu_w)$ and $(N_f, F_f, \mu_f)$, respectively, and the gravitational acceleration by $g$.

The first condition is expressed geometrically as \begin{equation} H_{\mathrm{lim}}^{\mathrm{step}} = L_b^{\mathrm{step}} + \frac{a}{\sqrt2}. \end{equation}

For the second condition, the friction coefficient required to prevent slipping is derived. When the CoM is located behind the step edge, the robot tends to rotate backward about the upper contact point; however, this rotation presses the lower body against the vertical wall, and the resulting wall normal force $N_w$ cancels the tip-over moment. From the moment balance about the upper contact point, $N_w$ is obtained as
\begin{equation}
N_w = \frac{\left( a + b + h_g - x_g^{\mathrm{step}} + y_g^{\mathrm{step}} \right) Mg}{\sqrt2\, L_b^{\mathrm{step}} + a - \sqrt2\,(h_g + r)}. 
\end{equation}

The wall friction force is related to the normal force by $F_w = \mu_w N_w$, and the vertical force equilibrium gives the upper-surface normal force
 \begin{equation}
 N_f = M g - \mu_w N_w. 
 \end{equation}

Here, $F_f = \mu_f N_f$, and the no-slip condition is $F_f \ge N_w$. Therefore, the second condition is expressed as 
\begin{equation}
\mu_f \left( Mg - \mu_w N_w \right) \ge N_w.
\end{equation}

Assuming, for simplicity, an equal friction coefficient for the wall and the upper surface ($\mu_w = \mu_f = \mu$), this condition reduces to 
\begin{equation}
N_w\, \mu^2 - Mg\, \mu + N_w \le 0.
\end{equation}

Substituting the design parameters in Table~\ref{tab:geometric_parameters} shows that slipping is prevented for $0.084 \le \mu \le 11.95$, which is readily satisfied under practical conditions. Hence, slipping does not limit the step traversal, and the maximum step height is governed by the first condition. In practice, the rear body length at the foremost position is set to approximately $L_b^{\mathrm{step}} \approx 680\,\mathrm{mm}$, yielding $H_{\mathrm{lim}}^{\mathrm{step}} = 708\,\mathrm{mm}$, corresponding to approximately $75.3\%$ of the robot body length.

This analysis demonstrates that, for a general vertical step, TRASER can traverse it stably as long as its front end can reach the upper surface, without being constrained by stability or slipping. Combined with the high front-end reachability established in the preceding analysis, this property leads to a remarkably high step-traversal capability.

\subsection{Suspended-Platform Traversal Model}
This subsection derives the theoretical maximum platform height, $H_{\mathrm{lim}}^{\mathrm{plat}}$, of TRASER based on a geometric condition. 

In the step traversal case, the wall reaction force cancels the tip-over moment, thereby ensuring static stability. In contrast, a suspended-platform has no vertical supporting surface, and thus no wall reaction force is available. Consequently, static stability at the instant when the rear end leaves the ground, corresponding to the transition from Fig.~\ref{tenban_str}(f) to (g), must be maintained solely by the CoM.

\begin{figure}[!t]
\centering
\includegraphics[width=0.90\linewidth]{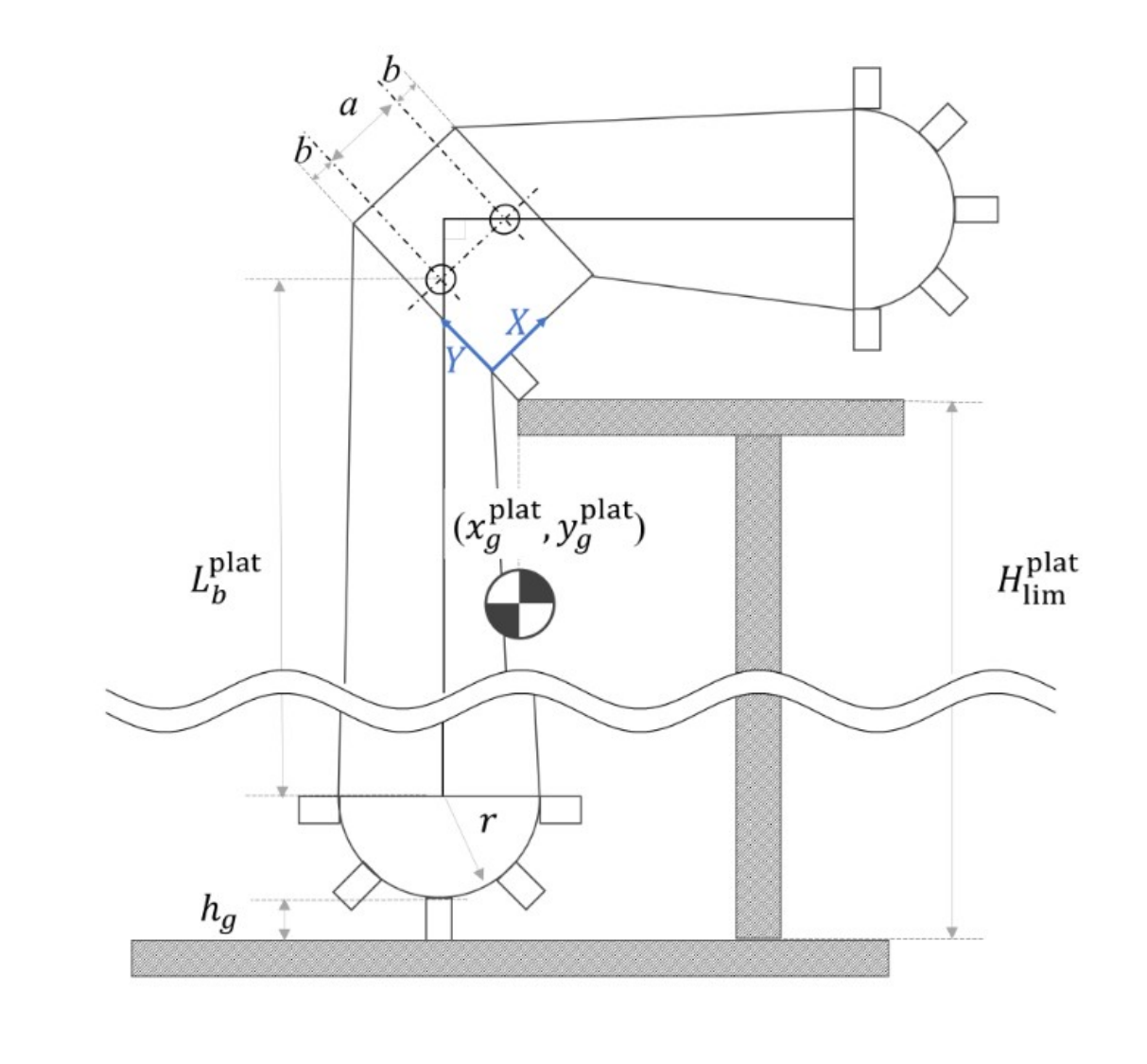}
\caption{Free-body diagram for suspended-platform traversal analysis. Without wall support, successful traversal requires the center of mass to move beyond the platform edge.}
\label{tenban_mdr}
\end{figure}

To evaluate this condition, the model shown in Fig.~\ref{tenban_mdr} is employed. In this model, the robot is assumed to be in a bent configuration ($\theta = -45^\circ$). Stability at this instant requires the CoM to be located in front of the platform edge along the traveling direction, giving the boundary condition \begin{equation} \frac{x_g^{\mathrm{plat}} - y_g^{\mathrm{plat}}}{\sqrt{2}} - \frac{h_g}{\sqrt{2}} = 0. \end{equation} Since $x_g^{\mathrm{plat}}$ and $y_g^{\mathrm{plat}}$ are both functions of the rear body length through (11) and (12), solving this boundary condition for the rear body length yields the value $L_b^{\mathrm{plat}}$ at the stability limit. Using this $L_b^{\mathrm{plat}}$, the theoretical maximum platform height is obtained geometrically as
\begin{equation}
H_{\mathrm{lim}}^{\mathrm{plat}} = L_b^{\mathrm{plat}} + r + h_g - \frac{1}{\sqrt{2}}\left(h_g + r + b\right). \end{equation}

Substituting the design parameters in Table~\ref{tab:geometric_parameters} yields a theoretical maximum platform height of $635\,\mathrm{mm}$, corresponding to approximately $67.6\%$ of the robot body length. Compared with step traversal, suspended-platform traversal yields a smaller traversal limit because the wall reaction force cannot be exploited; consequently, the traversal performance is governed by the mechanical capability of substantially shifting the center of mass forward.

\subsection{Ditch Traversal Model}

This subsection derives the theoretical maximum ditch width, $G_{\mathrm{lim}}$, of TRASER based on a geometric condition. The ditch traversal capability is an important performance index for robots operating in disaster environments, where cracks and collapsed structures are frequently encountered.

As illustrated in Fig.~\ref{beyond_str}, successful ditch traversal requires satisfying two geometric conditions. The first condition is that the robot must reach the opposite edge of the ditch, as shown in Fig.~\ref{beyond_str}(c)--(d), which is governed by the front-end reachability. The second condition is static stability during the transition shown in Fig.~\ref{beyond_str}(f)--(g), where the rear end loses contact with the near side of the ditch. At this instant, the robot is supported only by the opposite edge, and tip-over is avoided only if the CoM remains beyond the supporting edge. Since TRASER has sufficiently high front-end reachability, the maximum traversable ditch width is predominantly determined by the second stability condition.

\begin{figure}[!t]
\centering
\includegraphics[width=0.90\linewidth]{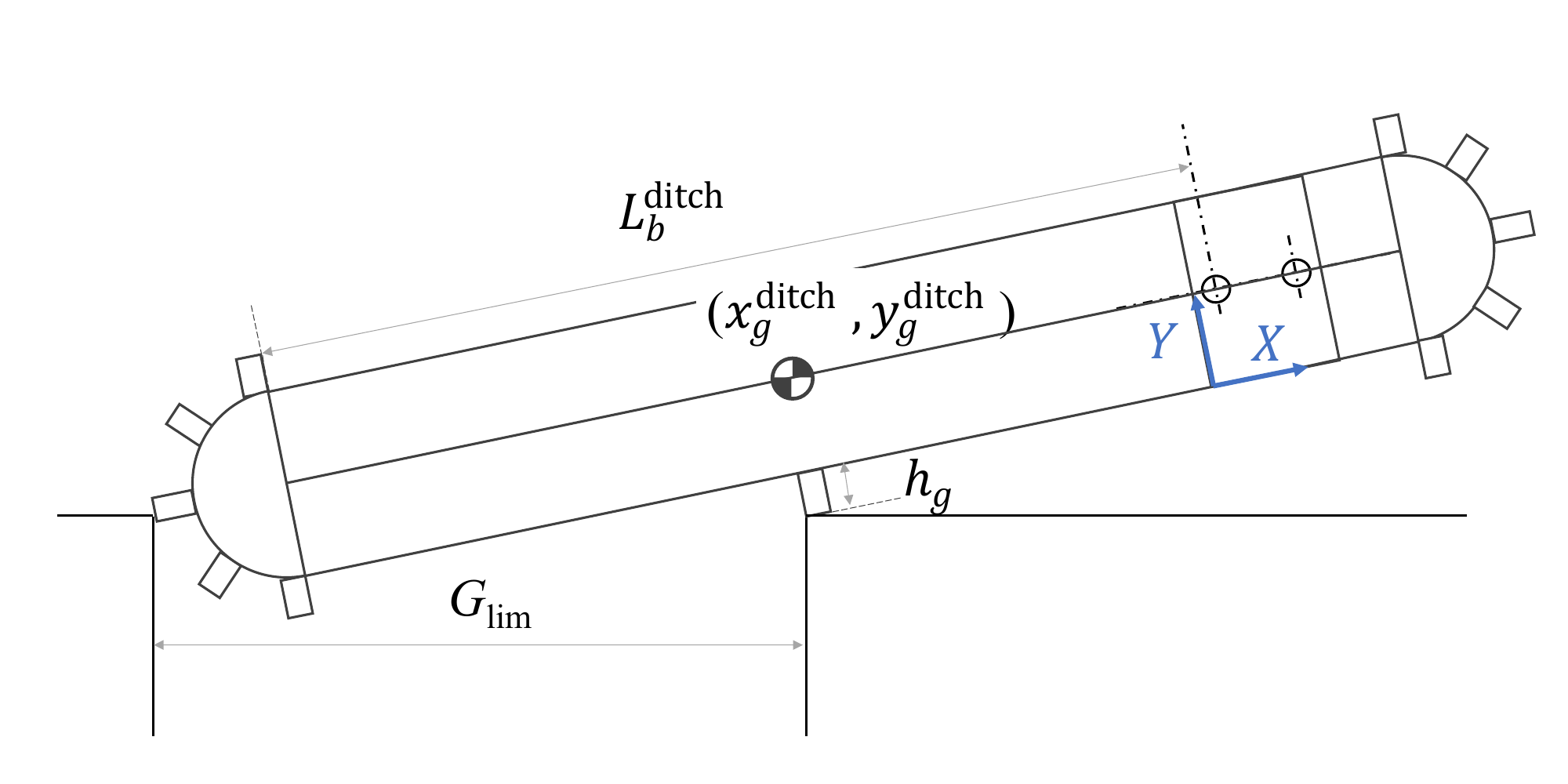}
\caption{Free-body diagram used to derive the maximum traversable ditch width.}
\label{beyond_mdr}
\end{figure}

To evaluate this condition, the model shown in Fig.~\ref{beyond_mdr} is employed. In this model, the robot is assumed to be in the straight configuration ($\theta = 0^\circ$) with the bending unit placed at its foremost position. The rear body length in this configuration is denoted by $L_b^{\mathrm{ditch}}$, and the CoM position is then given by (11) and (12) as $(x_g^{\mathrm{ditch}}, y_g^{\mathrm{ditch}})$. Applying the Pythagorean theorem to the geometry shown in Fig.~\ref{beyond_mdr}, the limiting ditch width is obtained as

\begin{equation} G_{\mathrm{lim}}= \sqrt{(L_b^{\mathrm{ditch}} + x_g^{\mathrm{ditch}} + r + h_g-b)^2 + (r+h_g)^2}. \end{equation}

In practice, the rear body length at the foremost position is set to approximately $L_b^{\mathrm{ditch}} \approx 700\,\mathrm{mm}$. Substituting this value together with the design parameters in Table~\ref{tab:geometric_parameters} yields a theoretical maximum ditch width of $579\,\mathrm{mm}$, corresponding to approximately $61.6\%$ of the robot body length.

\section{Experiment}
In this section, experimental environments equivalent to those used in the modeling of Section~IV were prepared. The obstacle traversal capability of TRASER was evaluated under manual operation. Detailed motion sequences are provided in the accompanying supplementary video. Throughout all experiments, neither belt derailment nor significant slippage occurred, confirming that the effect of the belt-path length variation is sufficiently small and supporting the validity of the belt-path variation model.

\subsection{Step Traversal Test}
\begin{figure}[!t]
\centering
\includegraphics[width=0.90\linewidth]{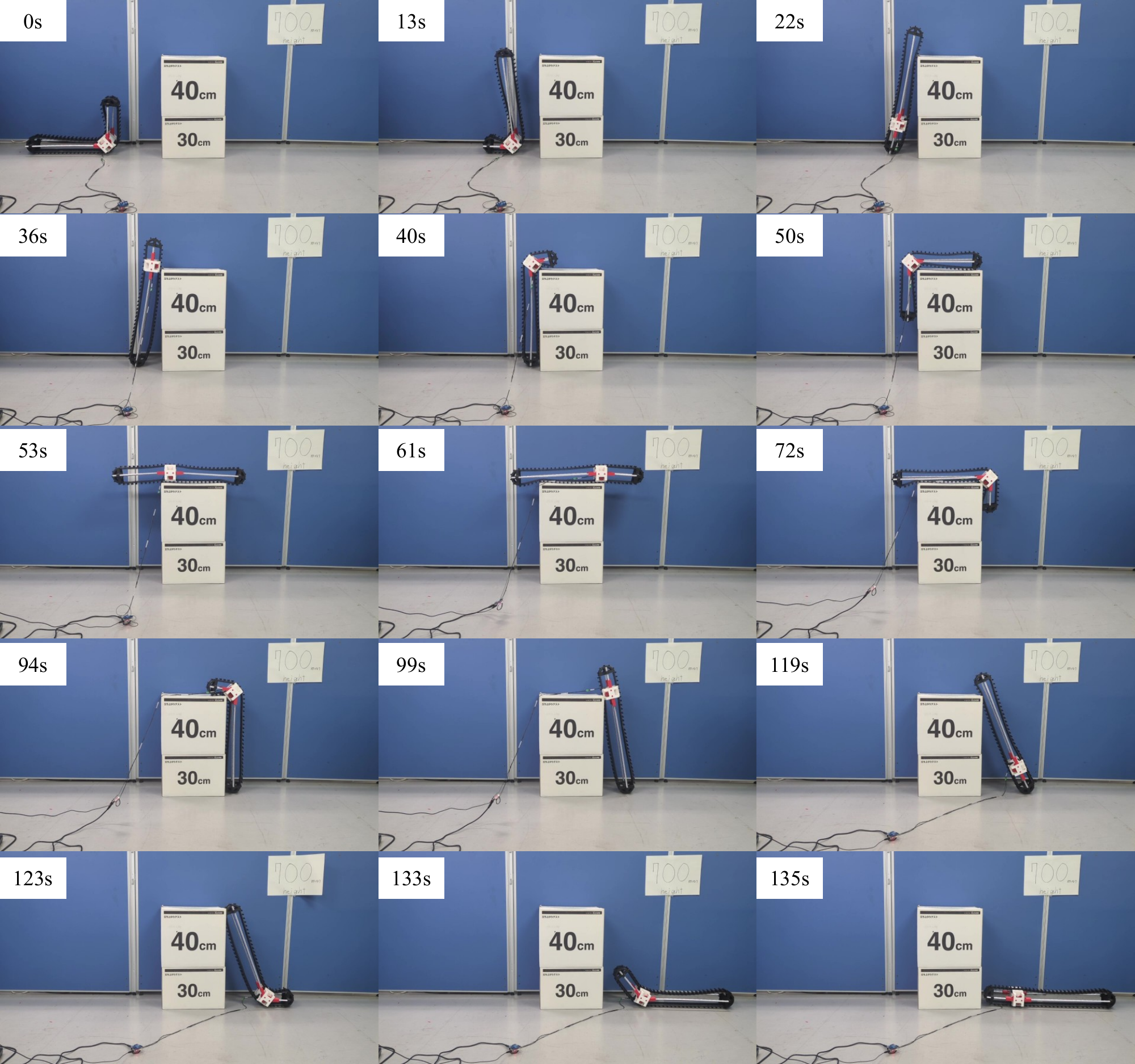}
\caption{Snapshots of the step traversal experiment, demonstrating traversal of a 700 mm step (74\% of the robot body length).}
\label{step}
\end{figure}

A vertical step was constructed by stacking cardboard boxes, as shown in Fig.~\ref{step}. The maximum step height was evaluated by increasing the obstacle height from $600\,\mathrm{mm}$ in $50\,\mathrm{mm}$ increments.

The robot first translated the bending unit toward the front of the body and bent the body to raise the front end above the step. After establishing contact with the upper surface, the bending unit was translated onto the step while the body articulated to maintain contact with both the wall and the upper surface. The robot then completed the climb and subsequently descended using the reverse sequence.

The robot successfully traversed a $700\,\mathrm{mm}$ step, corresponding to 74\% of its body length. This result closely agrees with the theoretical prediction of $H_{\mathrm{lim}}^{\mathrm{step}}=708\,\mathrm{mm}$, with an error of approximately $1.1\%$, validating the analytical model.

\subsection{Suspended-Platform Traversal Test}
\begin{figure}[!t]
\centering
\includegraphics[width=0.90\linewidth]{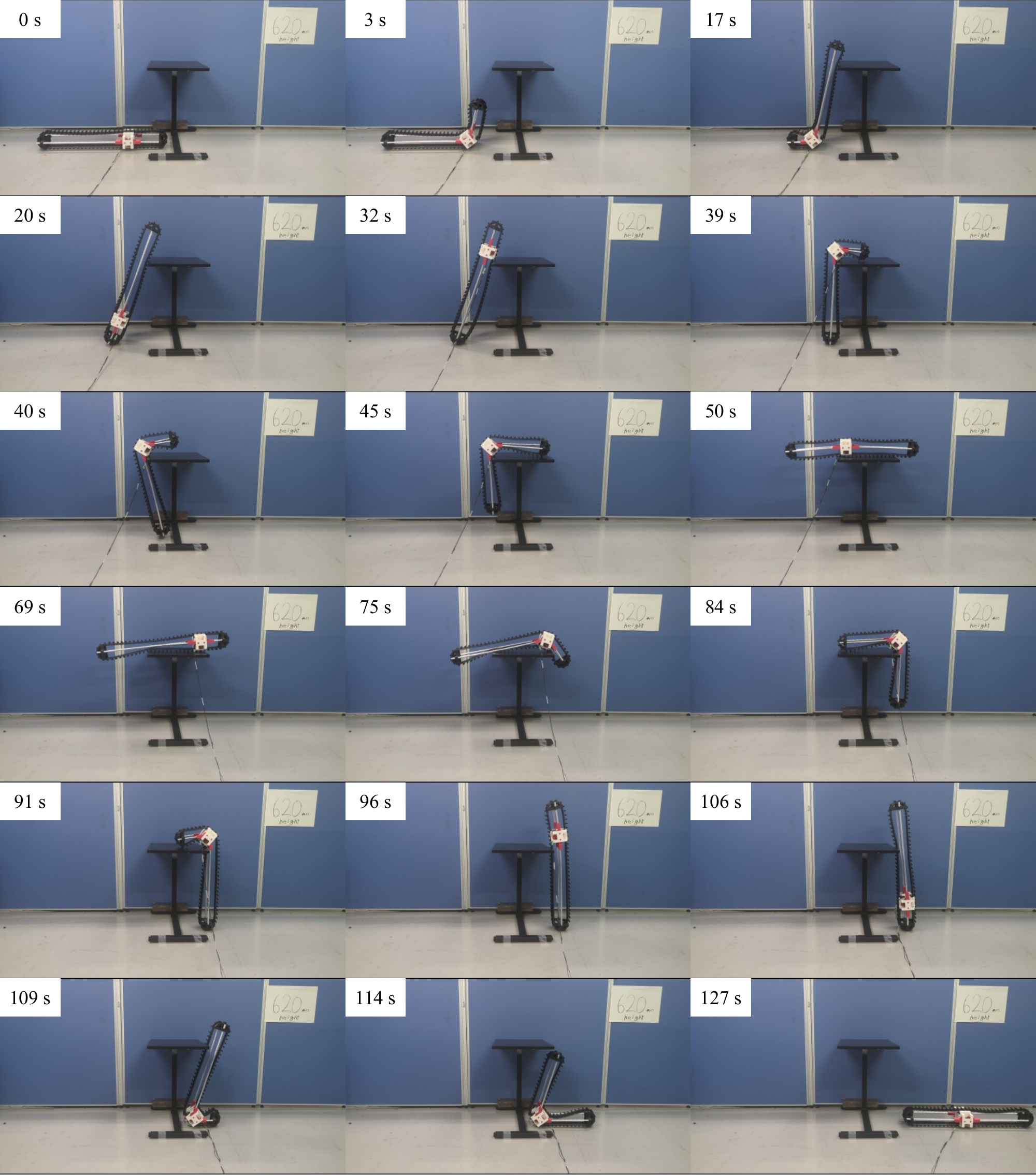}
\caption{Snapshots of the suspended-platform traversal experiment, demonstrating traversal of a 620 mm platform (66\% of the robot body length).}
\label{tenban}
\end{figure}

A suspended-platform without a vertical supporting surface was prepared, as shown in Fig.~\ref{tenban}. The platform height was increased in $50\,\mathrm{mm}$ increments to determine the maximum traversable height.

The robot approached the platform by lifting its front end without relying on wall reaction forces. After contacting the upper surface, the bending unit was translated forward to relocate the center of mass, allowing the robot to climb onto the platform while maintaining stability. The robot then descended using the reverse sequence.

TRASER successfully traversed a $620\,\mathrm{mm}$ platform (66\% of the body length), slightly below the theoretical value of $H_{\mathrm{lim}}^{\mathrm{plat}}=635\,\mathrm{mm}$, with an error of approximately $2.4\%$. This discrepancy is attributed to local track-belt deformation at the platform edge, which shifted the effective center of mass rearward relative to the analytical model.

\subsection{Ditch Traversal Test}
\begin{figure}[!t]
\centering
\includegraphics[width=0.90\linewidth]{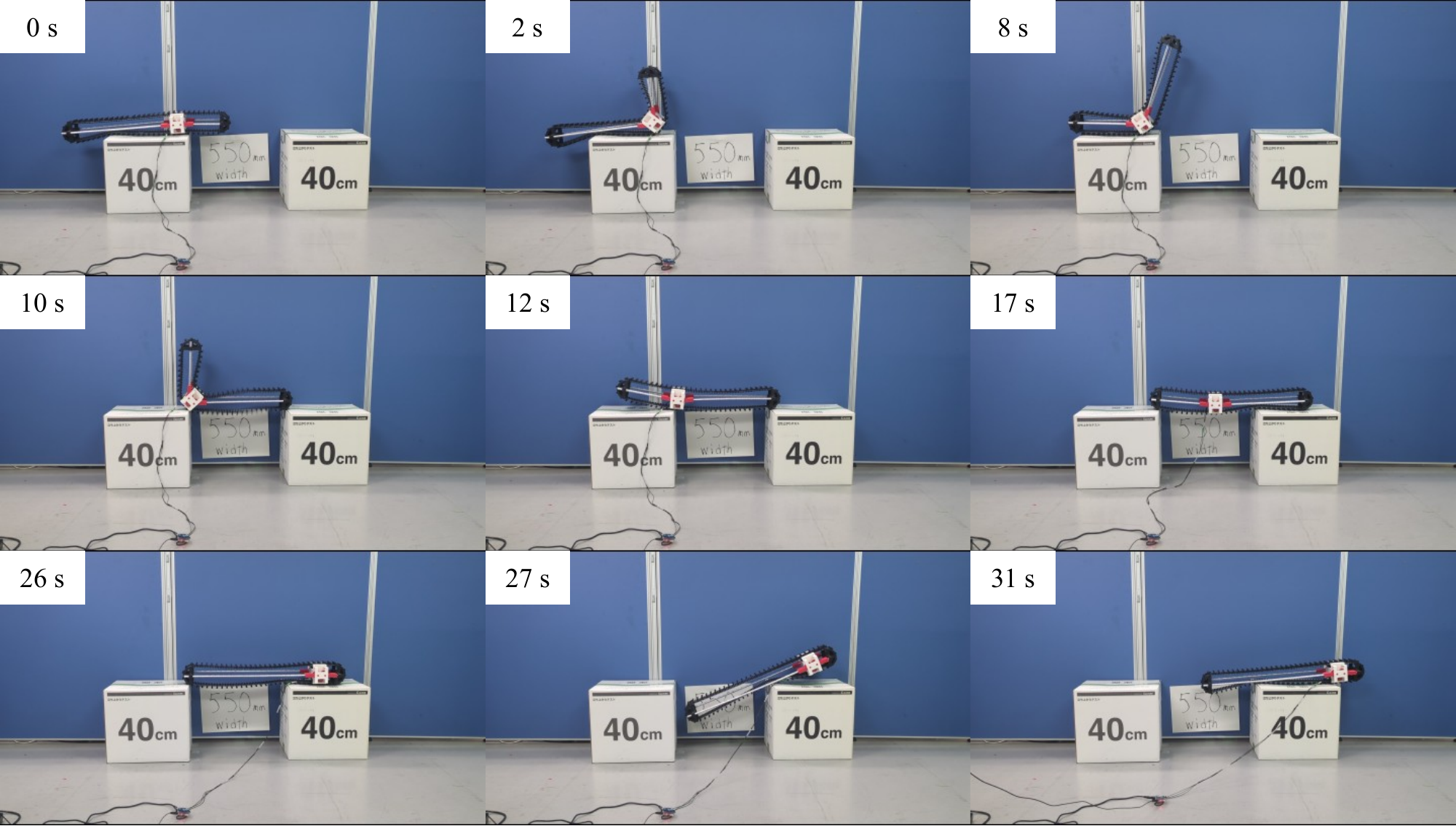}
\caption{Snapshots of the ditch traversal experiment, demonstrating traversal of a 550 mm ditch (59\% of the robot body length).}
\label{beyond}
\end{figure}

As shown in Fig.~\ref{beyond}, the ditch width was increased from $500\,\mathrm{mm}$ in $10\,\mathrm{mm}$ increments to determine the maximum traversable distance.

The robot first lifted its front end to reach the opposite edge while keeping the center of mass near the rear. After the front end established contact, the bending unit was translated forward to shift the center of mass onto the opposite side before the rear end left the ground, enabling stable ditch traversal.

The robot successfully crossed a $550\,\mathrm{mm}$ ditch (59\% of the body length), compared with the theoretical prediction of $G_{\mathrm{lim}}=579\,\mathrm{mm}$, with an error of approximately $5.0\%$. This discrepancy is mainly attributed to track-belt deformation at the opposite edge, which caused a slight rearward shift of the effective center of mass.

\subsection{Comparison with Existing Tracked Robots}
The traversal performances obtained in the previous experiments are summarized in Table~\ref{tab:tracked_robot_comparison} together with those of representative tracked robots. The compared robots represent a wide variety of designs, including sub-tracked~\cite{kenaf}, variable-geometry~\cite{Kinugasa2016}, articulated~\cite{souryu}, mass-shifting~\cite{app12010525}, tail-assisted~\cite{Seo2013FlipBot}, and reconfigurable types~\cite{8930917} and our previous tracked robot~\cite{uda2025crawler}. For a fair comparison, the step-climbing height and ditch traversal distance are normalized by the total robot length and expressed as fractions of the robot body length. For Kenaf and FlipBot, the robot lengths include the auxiliary sub-tracks and supporting tail, respectively, and are estimated from the published figures.

As shown in Table~\ref{tab:tracked_robot_comparison}, TRASER achieves normalized step traversal and ditch traversal ratios of 0.74 and 0.59, respectively, the highest among the compared robots. In particular, the normalized step traversal ratio substantially exceeds those of the other robots, which range from 0.26 to 0.46. This improvement is attributed to resolving the reachability--stability tradeoff inherent in existing tracked robots with fixed bending locations and mass distributions through the coordination of arbitrary-location articulation and substantial CoM relocation. These results demonstrate that TRASER provides superior obstacle traversal performance compared with existing tracked robots.

\begin{table*}[!t] \centering \caption{Comparison of Obstacle Traversal Performance Among Representative Tracked Robots} \label{tab:tracked_robot_comparison} \renewcommand{\arraystretch}{1.2} \begin{tabular}{l l c c c} \toprule Robot & Type & \begin{tabular}[c]{@{}c@{}}Width $\times$ Length $\times$ Height\\ (mm)\end{tabular} & \begin{tabular}[c]{@{}c@{}}Step traversal ability\\ (step height / robot length)\end{tabular} & \begin{tabular}[c]{@{}c@{}}Ditch traversal ability\\ (ditch distance / robot length)\end{tabular} \\ \midrule Kenaf~\cite{kenaf} & Sub-tracked & $(431 \times 941 \times 195)^{*}$ & 0.32 & 0.32 \\ RT04-NAGA~\cite{Kinugasa2016} & Variable-geometry & $220 \times 1250 \times 160$ & 0.36 & 0.53 \\ Souryu-V~\cite{souryu} & Articulated & $202 \times 1380 \times 145$ & 0.46 & --- \\ Fukuoka's robot~\cite{app12010525} & Mass-shifting & $600 \times 360 \times 180$ & 0.30 & 0.53 \\ FlipBot~\cite{Seo2013FlipBot} & Tail-assisted & $(480 \times 778 \times 160)^{*}$ & 0.26 & --- \\ Zarrouk's robot~\cite{8930917} & Reconfigurable & $120 \times 660 \times 104$ & 0.45 & 0.45 \\ Our previous robot~\cite{uda2025crawler} & Reconfigurable & $130 \times 470 \times 145$ & 0.44 & 0.54 \\ \midrule \textbf{TRASER} & \textbf{Reconfigurable} & $\mathbf{220 \times 940 \times 130}$ & \textbf{0.74} & \textbf{0.59} \\ \bottomrule \end{tabular} \\[3pt] {\footnotesize $^{*}$\,Estimated from the published figures.\quad ``---''\,indicates that the corresponding data were not reported in the literature.} \end{table*}

\section{Conclusions}
This paper presented TRASER, a reconfigurable tracked robot that combines a movable articulation point with large CoM shifting to improve obstacle traversal capability. Geometric and static analyses were developed to clarify the relationships among the articulation point, CoM position, and obstacle traversal performance. The analytical models also provided theoretical limits for step traversal, suspended-platform traversal, and ditch traversal.

Experiments validated the proposed analyses and demonstrated successful traversal of a 700 mm step, a 620 mm suspended-platform, and a 550 mm ditch, corresponding to 74\%, 66\%, and 59\% of the robot body length, respectively. These results demonstrate state-of-the-art obstacle traversal performance among tracked mobile robots and confirm the effectiveness of combining localized body bending with active CoM shifting.

Future work will focus on improving the maneuverability and payload capacity of TRASER. Since the current prototype provides only straight-line locomotion, steering mechanisms or cooperative locomotion using multiple robots will be investigated. In addition, because the tape-spring spine supports the entire robot body, its load-carrying capacity is limited by buckling under large external loads. Future work will therefore investigate structural reinforcement and load-support mechanisms to increase the payload while preserving the robot's reconfigurability.

\bibliographystyle{IEEEtran}
\bibliography{reference}

\begin{IEEEbiography}[{\includegraphics[width=1in,height=1.25in,clip,keepaspectratio]{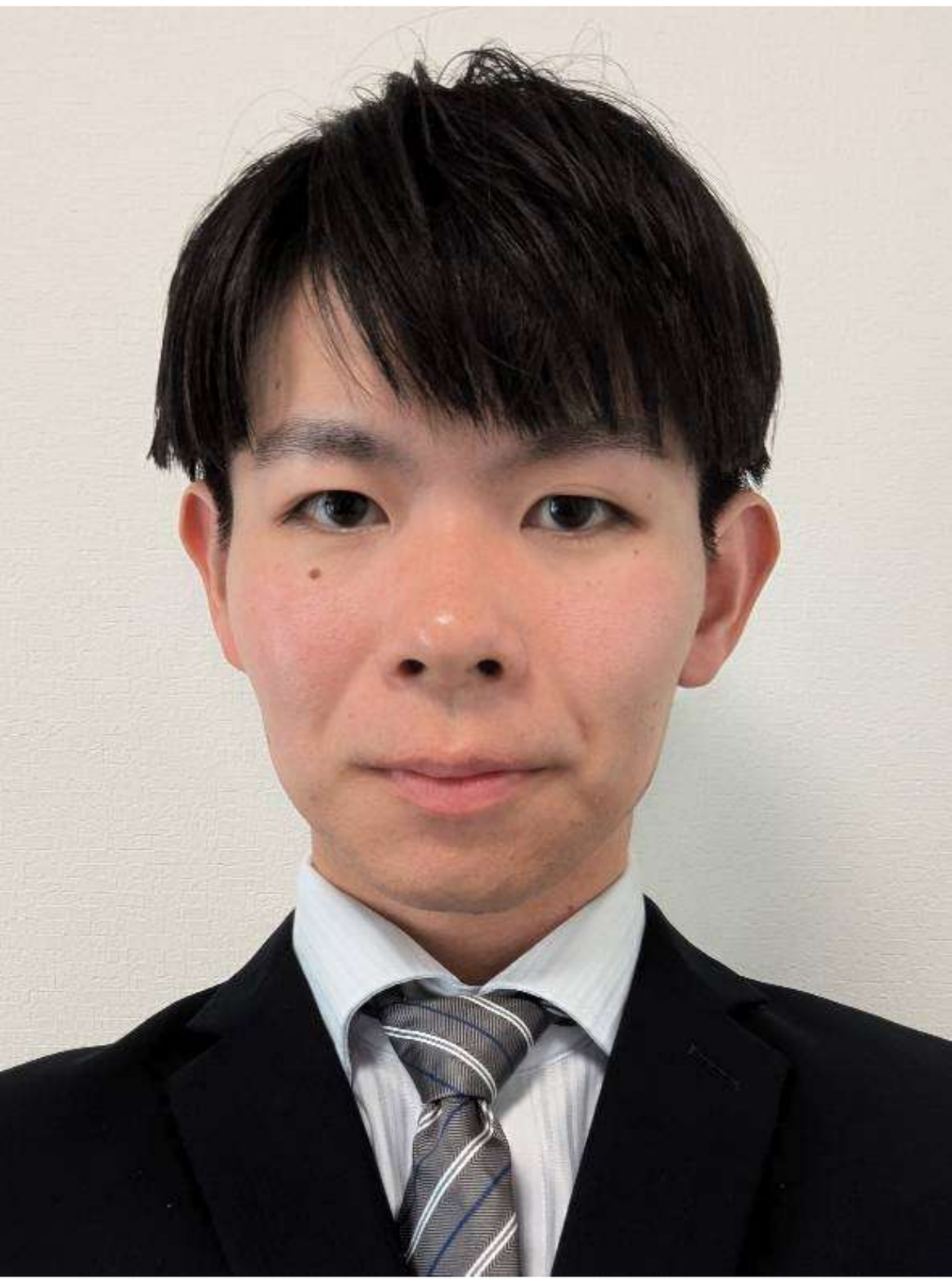}}]{Yuki Uda} (Student Member, IEEE) received the B.S. degree in mechanical engineering from Kyushu University, Fukuoka, Japan, in 2025, where he is currently pursuing the M.S. degree. His research interests include mobile robots and reconfigurable mechanisms. \end{IEEEbiography}

\begin{IEEEbiography}[{\includegraphics[width=1in,height=1.25in,clip,keepaspectratio]{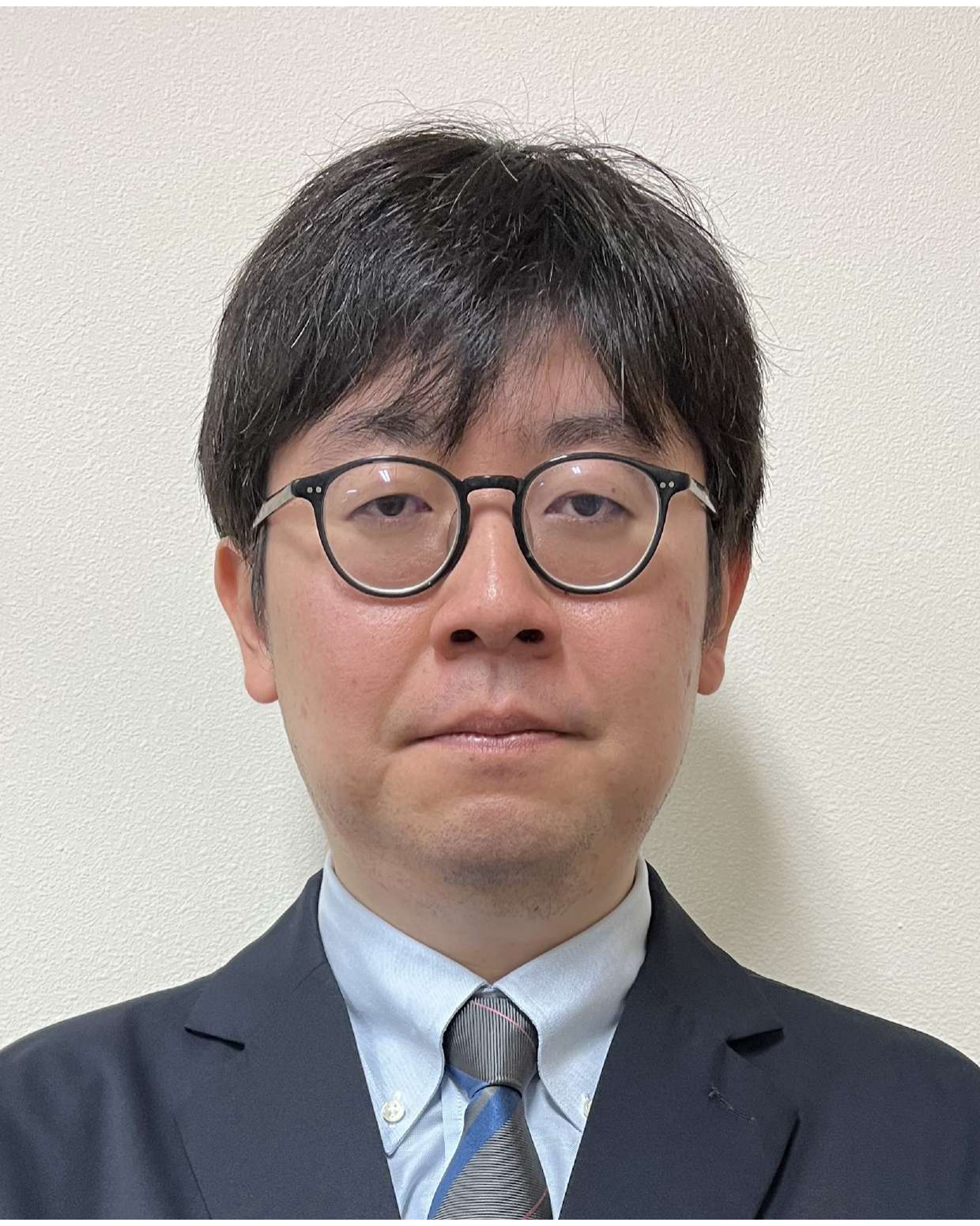}}]{Yasutaka Nakashima} (Member, IEEE) received the B.S. and M.S. degrees from Waseda University, Tokyo, Japan, in 2009 and 2011, respectively, and the Ph.D. degree in engineering from Waseda University in 2014. He was an Assistant Professor with the Faculty of Engineering, Kyushu University, Fukuoka, Japan, from 2014 to 2019, where he has been an Associate Professor since 2019. His research interest includes assistive robotics. \end{IEEEbiography}

\begin{IEEEbiography}[{\includegraphics[width=1in,height=1.25in,clip,keepaspectratio]{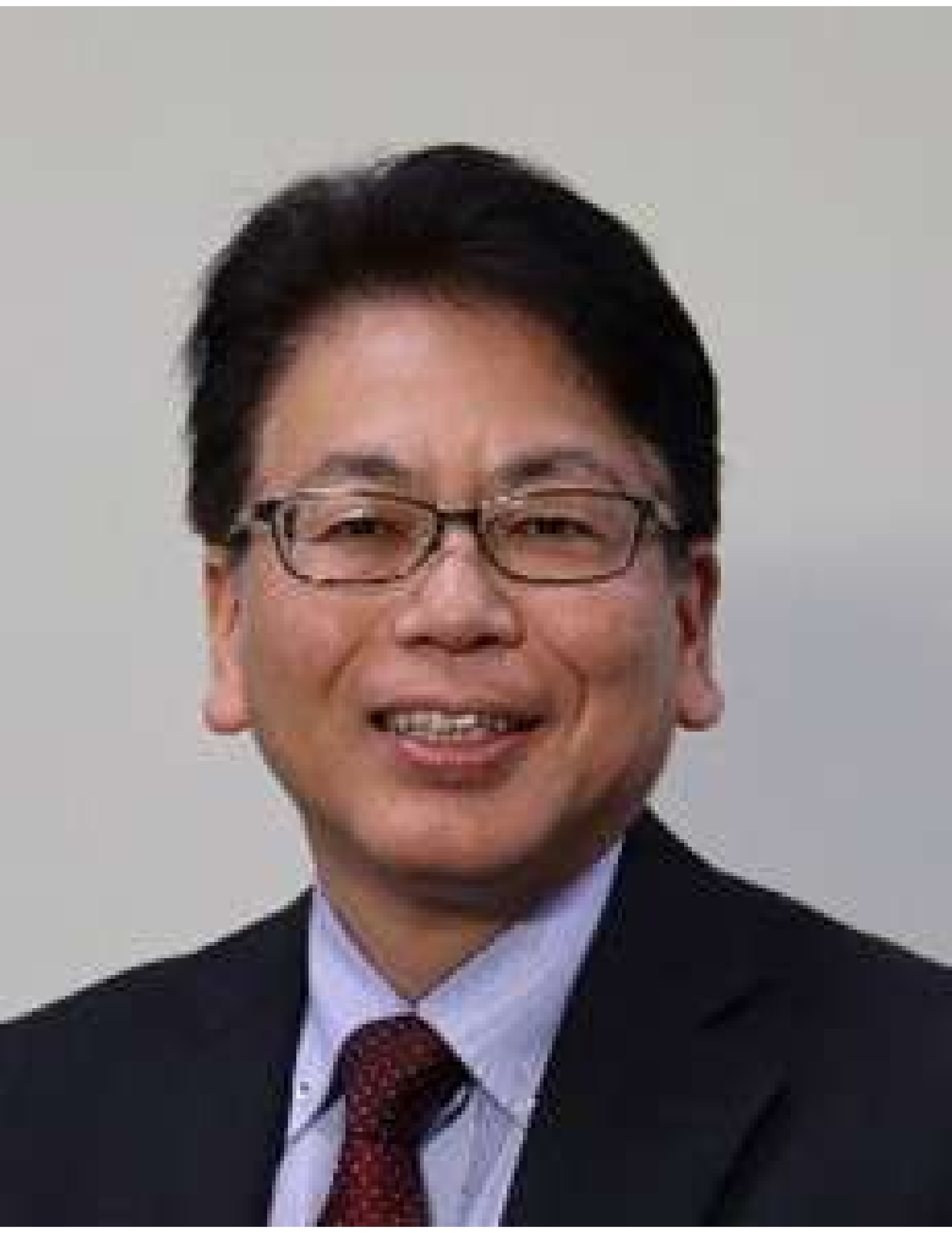}}]{Motoji Yamamoto} (Member, IEEE) received the B.E., M.E., and Doctor of Engineering degrees from Kyushu University, Fukuoka, Japan, in 1985, 1987, and 1990, respectively. He was a Lecturer and an Associate Professor with Kyushu University from 1990 to 1992 and from 1992 to 2005, respectively. Since 2005, he has been a Professor with the Department of Mechanical Engineering, Kyushu University. His research interests include wire suspended mechanisms, service robots, and human care robots. \end{IEEEbiography}

\begin{IEEEbiography}[{\includegraphics[width=1in,height=1.25in,clip,keepaspectratio]{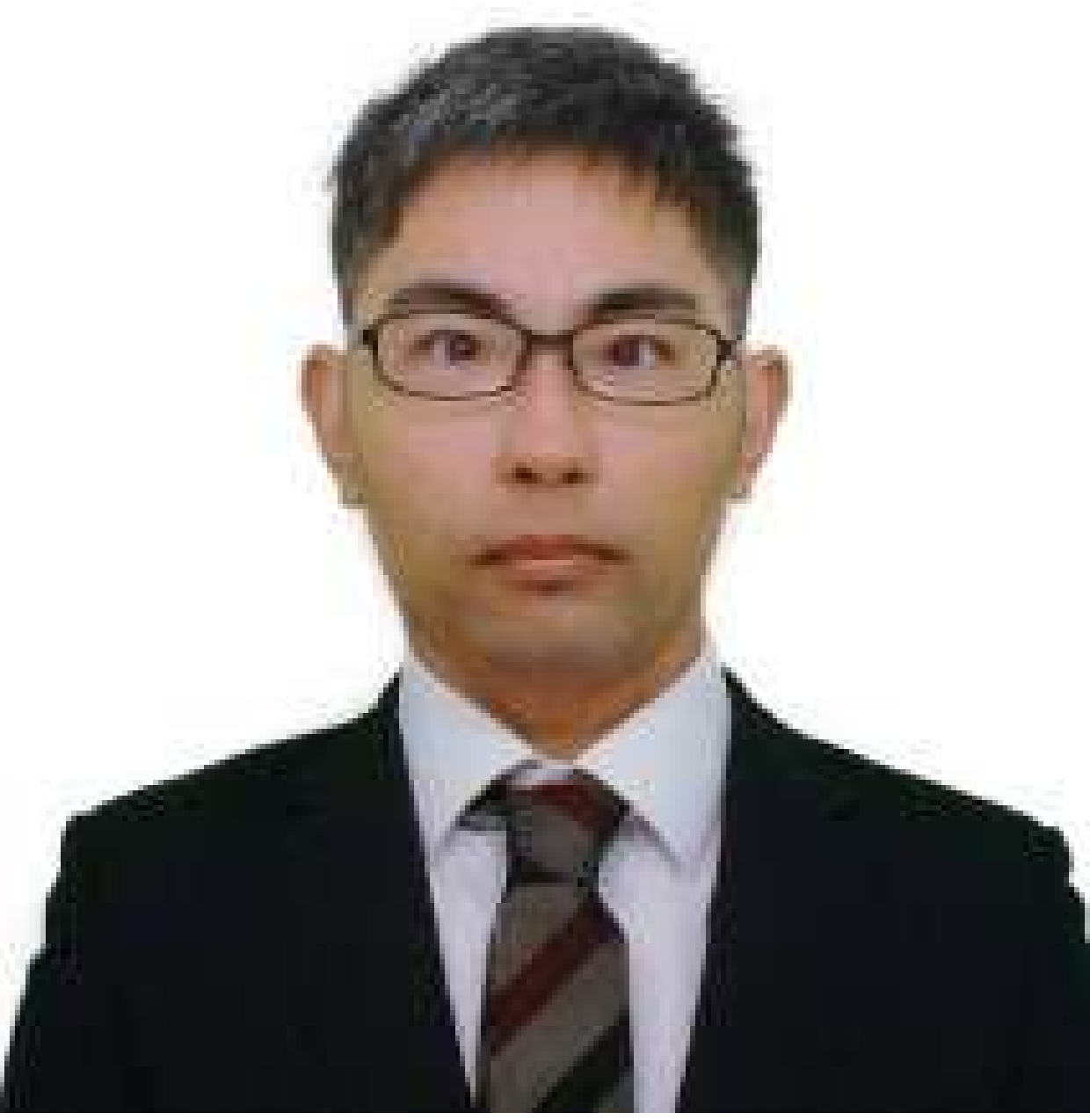}}]{Ayato Kanada} (Member, IEEE) received the Ph.D. degree in mechanical engineering from Toyohashi University of Technology, Aichi, Japan, in 2020. He was an Assistant Professor with the Department of Mechanical Engineering, Kyushu University, Fukuoka, Japan. In 2025, he became an Associate Professor with the Graduate School of Informatics and Engineering, The University of Electro-Communications, Tokyo, Japan. \end{IEEEbiography}

\end{document}